\documentclass[10pt,lineno,dvipsnames,sort]{customstyle}
\usepackage{doi}
\usepackage{makecell} 

\usepackage{helvet}
\usepackage{wrapfig}
\usepackage{floatrow} 
\usepackage{comment} 

\addto\extrasenglish{%

}

\usepackage{amsmath}
\usepackage{dsfont}
\newcommand{\mathbold}[1]{\ensuremath{\boldsymbol{\mathbf{#1}}}}

\usepackage[algoruled,noend]{algorithm2e}
\usepackage{listings}
\usepackage{fancyvrb}
\fvset{fontsize=\small}

\renewcommand{\d}[1]{\ensuremath{\operatorname{d}\!{#1}}}

\newcommand{\bx}{\mathbold{x}}

\newcommand{\btheta}{\mathbold{\theta}}

\title{Simulation-Based Inference: A Practical Guide}

\author[1,2]{Michael~Deistler*}
\author[1,2,3]{Jan~Boelts*} 
\author[4]{Peter~Steinbach} 
\author[1,2]{Guy~Moss} 
\author[5]{Thomas~Moreau} 
\author[1,2]{Manuel~Gloeckler}
\author[6]{Pedro~L.~C.~Rodrigues}
\author[5]{Julia~Linhart} 
\author[1,2]{Janne~K.~Lappalainen} %
\author[7]{Benjamin~Kurt~Miller} 
\author[1,2,8,9]{Pedro~J.~Gonçalves} 
\author[10]{Jan-Matthis~Lueckmann}  
\author[1,2]{Cornelius~Schröder} 
\author[1,2,11]{Jakob~H.~Macke} 

\affil[1]{Machine Learning in Science, University of Tübingen}
\affil[2]{Tübingen AI Center, Tübingen, Germany}
\affil[3]{TransferLab, appliedAI Institute for Europe}
\affil[4]{Helmholtz AI and Helmholtz-Zentrum Dresden-Rossendorf, Dresden, Germany}
\affil[5]{Université Paris-Saclay, Inria, CEA, Palaiseau, 91120, France}
\affil[6]{Univ.~Grenoble Alpes, Inria, CNRS, Grenoble INP, LJK}
\affil[7]{University of Amsterdam}
\affil[8]{VIB-Neuroelectronics Research Flanders (NERF), Belgium}
\affil[9]{Departments of Computer Science and Electrical Engineering, KU Leuven, Belgium}
\affil[10]{Google Research}
\affil[11]{Department Empirical Inference, Max Planck Institute for Intelligent Systems, Tübingen, Germany}

\corr{michael.deistler@uni-tuebingen.de}{}
\corr{jan.boelts@mailbox.org}{}
\corr{jakob.macke@uni-tuebingen.de}{}

\contrib[*]{These authors contributed equally to this work}

\begin{document}
\maketitle

\begin{abstract}
A central challenge in many areas of science and engineering is to identify model parameters that are consistent with prior knowledge and empirical data. Bayesian inference offers a principled framework for this task, but can be computationally prohibitive when models are defined by stochastic simulators. Simulation-based Inference (SBI) is a suite of methods developed to overcome this limitation, which has enabled scientific discoveries in fields such as particle physics, astrophysics, and neuroscience. 
The core idea of SBI is to train neural networks on data generated by a simulator, without requiring access to likelihood evaluations. Once trained, inference is amortized: The neural network can rapidly perform Bayesian inference on empirical observations without requiring additional training or simulations. 
In this tutorial, we provide a practical guide for practitioners aiming to apply SBI methods.
We outline a structured SBI workflow and offer practical guidelines and diagnostic tools for every stage of the process---from setting up the simulator and prior, choosing and training inference networks, to performing inference and validating the results.
We illustrate these steps through examples from astrophysics, psychophysics, and neuroscience.
This tutorial empowers researchers to apply state-of-the-art SBI methods, facilitating efficient parameter inference for scientific discovery.
\end{abstract}

\section{Introduction}
\label{sec:1_introduction}

\begin{figure}[th]
    \centering
    \includegraphics[width=1\textwidth]{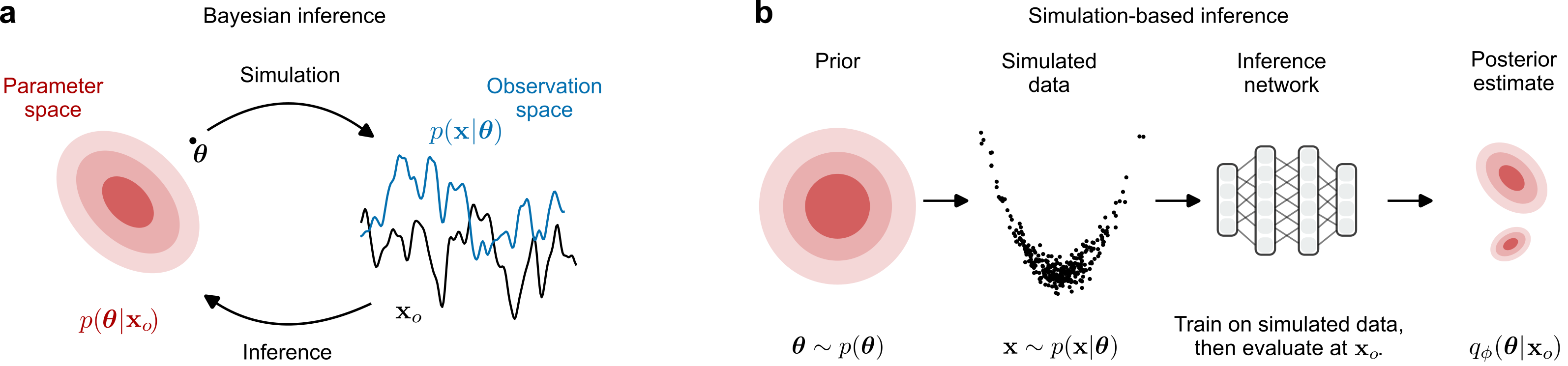}
    \caption{\textbf{Bayesian inference and simulation-based inference (SBI).}
    \textbf{(a)} Bayesian inference solves an inverse problem: Given a stochastic simulator which takes in parameters $\btheta$ and returns simulations $\bx \sim p(\bx | \btheta)$, it infers parameters which could have generated observed data $\bx_o$ and are consistent with prior knowledge.
    \textbf{(b)} Simulation-based inference (SBI) encompasses a range of algorithms that perform this statistical inference task. Neural network-based methods for SBI consist of four main steps:
    (i) First, we draw parameters from the prior distribution $\btheta \sim p(\btheta)$, and (ii) run the simulator for each of these parameters to generate simulated data $\bx$.
    (iii) We then train an inference network $q_\phi$ on these simulated data. In neural posterior estimation (NPE), this inference network is a conditional generative model that takes as input $\bx$ and predicts a distribution of parameters.
    (iv) After training, we evaluate the inference network at the observation $\bx_o$ to obtain an approximation to the posterior $p(\btheta|\bx_o)$.
    }
    \label{fig:1_overview}
\end{figure}


Many fields in science and engineering rely on computer simulations to understand the mechanisms underlying observed phenomena.
These simulators serve as tools to generate hypotheses, design experiments, and uncover causal relationships.
However, many of these simulators are stochastic and include parameters that cannot be directly measured or are prohibitively expensive to estimate experimentally.
Inferring these parameters from data is essential, both to produce simulations matching the observed data as well as for scientific investigations. 

Bayesian inference has become a cornerstone of statistical analysis, offering a principled framework to quantify uncertainty in inferred parameters \citep{gelman1995bayesian}.
At its core, Bayesian inference seeks to recover the posterior distribution $p(\btheta | \bx_o)$, which represents the probability of a set of parameters $\btheta$ given observed data $\bx_o$ (Fig.~\ref{fig:1_overview}a).
This posterior provides a comprehensive view of the entire space of parameters that are compatible with the data and the prior, and thus contains information about parameter uncertainty, parameter dependencies, and plausible solutions.
However, when applied to simulators, classical Bayesian methods such as Markov chain Monte Carlo (MCMC) \cite{gilks1995markov} or variational inference (VI) \cite{zhang2018advances} often face significant challenges.
These methods typically rely on explicit likelihood evaluations, which are computationally expensive or even infeasible for many simulators \cite{csillery2010approximate}.
In particular, simulators are sometimes black-box programs that can generate data but do not provide closed-form likelihood functions or gradients with respect to their parameters \citep{brehmer2022simulation, gonccalves2020training, beaumont2002approximate}.

Simulation-based inference, also sometimes referred to as likelihood-free inference, has emerged as a powerful set of tools to perform Bayesian inference for these simulators \cite{cranmer2020frontier}. Recent advances build upon approaches such as Approximate Bayesian Computation \cite[ABC,][]{sisson_overview_2018}, and use modern tools from deep learning and generative modeling to enable more scalable posterior inference.
The key idea behind modern approaches in SBI is to first run the simulator to generate a dataset of parameters $\btheta$ and simulation outputs $\bx$, and to then train a neural network (called an inference network) to learn the probabilistic relationship between simulation outputs and parameters \cite{papamakarios2016fast,lueckmann2017flexible,greenberg2019automatic,papamakarios2019sequential,hermans2020likelihood}. After training, the neural network is evaluated at observed data $\bx_o$ to infer the posterior (Fig.~\ref{fig:1_overview}b).
By relying only on model simulations, SBI methods do not require explicit evaluations of likelihoods or gradients, making them applicable to any simulator. In addition, SBI methods treat the simulator as a ``black-box'', which allows simulations to be run fully in parallel and on any compute architecture or in any programming language. Finally, many SBI methods enable amortized inference: Once the neural network is trained, it can be used to rapidly perform inference for new observations without needing additional simulations or training, thus allowing real-time and high-throughput applications of Bayesian inference \cite{hermans2020likelihood,gonccalves2020training,radev2020bayesflow,dax2021real}. 

The goal of this tutorial is to provide an accessible and self-contained introduction to state-of-the-art SBI methods, guiding practitioners through the workflow of applying SBI to their own research problems.
The remainder of this paper is structured as a practical guide to the SBI workflow: 
We begin with an introductory example that illustrates the core concepts of Bayesian inference and SBI using a simple simulator (Section~\ref{sec:2_toy_example}).
Section~\ref{sec:3_workflow} then details the SBI workflow, providing guidelines and highlighting common pitfalls at each stage, from setting up the simulator and prior, choosing and training inference models, to performing inference and validating the results.
Finally, Section~\ref{sec:4_examples} demonstrates the application of the complete SBI workflow on three examples from astrophysics, psychophysics, and neuroscience.
We include code for all figures and examples, based on the widely used \texttt{sbi} toolbox \cite{boelts2024sbi}, to make it easy for readers to explore and build on these examples. 
We hope that this paper serves as a stepping stone for enabling robust scientific discovery with simulation-based inference for scientists in different disciplines.

\section{An introductory example}
\label{sec:2_toy_example}

\begin{figure}[t!]
    \centering
    \includegraphics[width=1\textwidth]{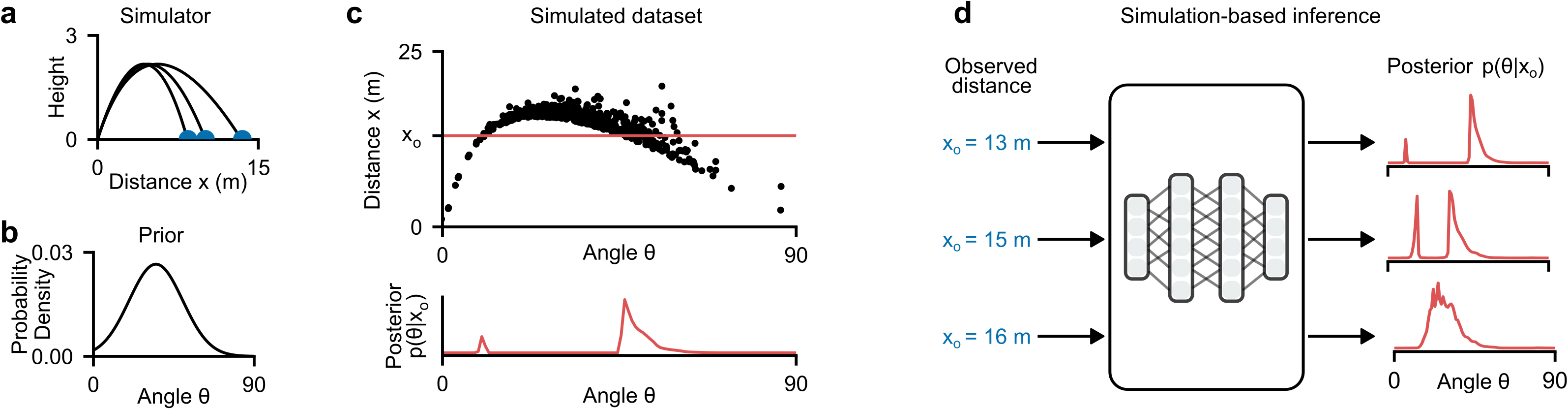}
    \caption{\textbf{Introductory example: SBI for a ball throwing simulator.}
    \textbf{(a)} Trajectories of the ball-throwing simulator. Only the final distance (blue) is used for inference.
    \textbf{(b)} Prior distribution for the throwing angle $\btheta$.
    \textbf{(c)} Top: Simulated dataset of $N$ throws, showing pairs of $(\btheta, \bx)$ samples from the joint distribution $p(\btheta, \bx)$ and a hypothetical observed distance $\bx_o=13$\,m (red line). Bottom: Posterior distribution $p(\btheta|\bx_o)$, i.e., the conditional distribution of parameters $\btheta$ given the observed distance $\bx_o$, visualized as a horizontal ``cut'' through the joint distribution. Note the bimodal shape, reflecting that $\bx_o=13$\,m could be produced both by low and high (but not by intermediate) angles.
    \textbf{(d)} Amortized SBI. An inference network is trained on simulated data to learn the mapping from an input distance $\bx$ to the corresponding posterior distribution $p(\btheta|\bx)$. Once it is trained, it can return the posterior for new observed distances $\bx_o$ without needing further simulations or retraining. Examples for three values of $\bx_o$ are shown.
    }
    \label{fig:2_ball_throw}
\end{figure}

We start by providing intuition for Bayesian inference in the context of simulators, as well as for the methodology to train neural networks to perform inference, using a simple example---simulation of a ball-throw. Given the angle at which it was thrown $\btheta$, the simulator models the distance a ball reaches $\bx$. In our simulator, the ball is always thrown with a fixed force, but we include tailwind as a noise source, and we assume an additional measurement noise in the recorded distance. Thus, the simulator is stochastic: Even if the ball is thrown multiple times at the exact same angle (input parameter $\btheta$), it will reach different distances $\bx$ (Fig.~\ref{fig:2_ball_throw}a). 

\subsection{Bayesian inference}
\label{sec:2_1_bayesian_inference}

\emph{Bayesian inference} solves an inverse problem: Given a measured distance $\bx_o$, it identifies angles $\btheta$ that could have produced this distance, taking into account prior knowledge of throwing angles. Since the simulator is stochastic, a single distance $\bx_o$ can potentially result from different angles---for example, a low angle with strong tailwind might produce the same distance as a higher angle with no wind. Bayesian inference provides a framework for quantifying this uncertainty by determining the probability distribution over possible angles $\btheta$ that could have generated the observed distance $\bx_o$. This distribution is known as the posterior distribution $p(\btheta | \bx_o)$.

To compute the posterior, Bayesian inference combines two ingredients: The \emph{prior distribution} $p(\btheta)$ and the \emph{likelihood} $p(\bx|\btheta)$. The prior distribution $p(\btheta)$ incorporates knowledge or beliefs about the possible values of the throwing angle $\btheta$ before observing any data, and can be based on expert knowledge or information collected via literature research. In our example, we use a Normal distribution  restricted to plausible angles between 0 and 90 degrees (`a Truncated Normal distribution'; Fig.~\ref{fig:2_ball_throw}b).
The likelihood $p(\bx|\btheta)$ is the conditional probability density of observing a specific distance $\bx$ given that the ball was thrown at a particular angle $\btheta$. It is defined by the simulator itself, as well as by the noise model (in this case, tailwind and measurement noise). For many complex simulators, directly evaluating this likelihood function $p(\bx|\btheta)$ can be challenging or impossible\footnote{For this simple ball-throw example, the likelihood could technically be calculated by integrating out the wind strength, thereby permitting likelihood evaluations \citep{gelman1995bayesian}. However, this is not the case for many real-world simulators that might have thousands of such latent variables.}. Importantly, we can easily \emph{sample} from the likelihood by running the simulator: For a given angle $\btheta$, running the simulator once provides a sample $\bx \sim p(\bx|\btheta)$.

The goal of Bayesian inference is to obtain the \emph{posterior distribution} $p(\btheta|\bx_o)$. This distribution represents the updated probability of different angles $\btheta$ after observing the distance $\bx_o$, combining information from both the likelihood and the prior via Bayes' theorem: $p(\btheta|\bx_o) \propto p(\bx_o|\btheta) p(\btheta)$.

The key idea of SBI is that we can estimate the posterior distribution from model simulations, without explicitly having to evaluate the likelihood function or using Bayes' rule: For example, assume that the measured distance was $\bx_o = 13$\,m, and we aim to recover the corresponding posterior distribution over throwing angles. To obtain it, SBI methods generate many pairs of $(\btheta, \bx)$ by first sampling angles $\btheta$ from the prior $p(\btheta)$ and then running the simulator with each of those angles to get distances $\bx \sim p(\bx|\btheta)$. This yields a dataset of $N$ samples $\{(\btheta_i,\bx_i)\}_{i=1,...,N}$ drawn from the joint distribution $p(\btheta, \bx) = p(\btheta)p(\bx|\btheta)$ (Fig.~\ref{fig:2_ball_throw}c, upper panel). The posterior distribution $p(\btheta|\bx_o)$ for a specific observed distance $\bx_o$ corresponds to the conditional distribution of $\btheta$ when fixing the data to $\bx_o$. Conceptually, this can be thought of as looking at a ``cut'' through the joint density $p(\btheta, \bx)$ at $\bx_o$ and inspecting the distribution of corresponding $\btheta$ values. For an observed distance of $13$\,m, the posterior distribution is bimodal (Fig.~\ref{fig:2_ball_throw}c, lower panel), revealing that this distance is consistent with angles in two distinct ranges.

\subsection{Simulation-based inference}
\label{sec:2_2_simulation_based_inference}

In practice, however, obtaining the posterior approximation from such a ``cut'' poses a fundamental challenge: For continuous data, no simulation will \emph{exactly} match the observed value $\bx_o$. Approximate Bayesian Computation (ABC) approaches address this by accepting parameters $\btheta$ whose simulations fall within some tolerance $\epsilon$ of the observation to approximate the posterior \cite{sisson_overview_2018}. This creates an inherent trade-off: Small tolerances ($\epsilon \rightarrow 0$) yield accurate posteriors but accept very few samples, while larger tolerances accept more samples but produce less accurate approximations. ABC methods face an additional challenge: They require defining a distance metric to measure how ``close'' a simulation $\bx_i$ is to the observation $\bx_o$. While straightforward for low-dimensional data, like our single distance measurement, designing effective distance metrics becomes increasingly difficult for higher-dimensional data. 

Recent SBI methods leverage neural networks to overcome these limitations: Rather than accepting or rejecting samples based on distance metrics, \emph{neural} SBI methods train neural networks to learn the relationship between simulated data $\bx$ and respective parameters $\btheta$ \cite{BlumFrancois_10,papamakarios2016fast}. We call these neural networks \emph{inference networks}. Once trained, these inference networks can be evaluated at any given observation to efficiently infer the posterior distribution over parameters. In this ball-throw example, we focus on \emph{Neural Posterior Estimation (NPE)}, a widely-used and conceptually simple SBI method \citep{papamakarios2016fast, lueckmann2017flexible, greenberg2019automatic, radev2020bayesflow}.

NPE trains an inference network to directly approximate the posterior distribution $p(\btheta|\bx)$, given samples $(\btheta, \bx)$ drawn from the joint distribution $p(\btheta, \bx)$. 
Instead of training the inference network to predict just a single best parameter value (a point estimate), NPE trains it to predict the parameters of a probability distribution over $\btheta$, conditioned on the input data $\bx$. We refer to this inference network as $q_\phi(\btheta | \bx)$. For example, assuming the posterior $p(\btheta | \bx)$ is approximately Gaussian, the network might learn a (nonlinear) mapping from an input-data $\bx$ to the mean and variance of a Gaussian distribution over the simulator parameters $\btheta$. The network $q_\phi$ is trained by minimizing the loss function
\begin{equation}
\label{eq:npe_loss}
   \mathcal{L}(\phi) =  \mathbb{E}_{(\btheta,\bx)\sim p(\btheta,\bx)}[-\log q_\phi (\btheta|\bx)],
\end{equation}
which encourages the network to assign high probability to the parameters $\btheta$ from the training data set that generated the corresponding (simulated) data $\bx$.
Crucially, this training process uses only simulated data. In practice, one often uses flexible generative models, such as normalizing flows \cite{papamakarios2021normalizing}, which can capture non-Gaussian posteriors\footnote{Normalizing flows are a type of neural network architecture particularly well-suited for modeling complex probability distributions, allowing both efficient sampling and efficient density evaluation.}.

Once the inference network is trained, it can be used to perform inference for any new observed data $\bx_o$. The trained network takes $\bx_o$ as input and directly provides the approximate posterior distribution $p(\btheta|\bx_o)$. A key advantage of this approach is that the network, once trained, can perform inference efficiently for multiple new observations without needing additional simulations or retraining (Fig.~\ref{fig:2_ball_throw}d). This property, known as \emph{amortized inference}, makes neural SBI powerful for analyzing many different observations from the same model. Amortization is particularly important for time-critical applications that are not accessible with classical Bayesian methods, like MCMC sampling \cite{dax2021real}, which often require time-consuming, iterative computations at test time.

We demonstrate the NPE workflow on the ball throwing simulator: First, we sample 1000 parameter sets from the prior and, for each of them, run the simulator. This generates a simulated dataset of 1000 $(\btheta, \bx)$ pairs. Second, we train a neural network on this dataset. This inference network (in this case, a normalizing flow) takes as input the simulated distance and predicts a distribution of angles. After training, we evaluate the inference network at three different observed distances ($13$\,m, $15$\,m, and $16$\,m). The inference network provides estimates for the posterior distribution given each distance with one forward pass (Fig.~\ref{fig:2_ball_throw}d).
We will discuss how to evaluate the correctness of such inference results in Section~\ref{sec:3_4_diagnostics}.

This ball-throwing example demonstrates how modern SBI overcomes the key limitation of traditional Bayesian methods: It performs inference without requiring explicit likelihood functions, using only simulated data to train neural networks. With this foundation, we now turn to a practical guide for using SBI.


\section{The SBI workflow: A practical step-by-step guide}
\label{sec:3_workflow}

\begin{figure}[t]
    \centering
    \includegraphics[width=\textwidth]{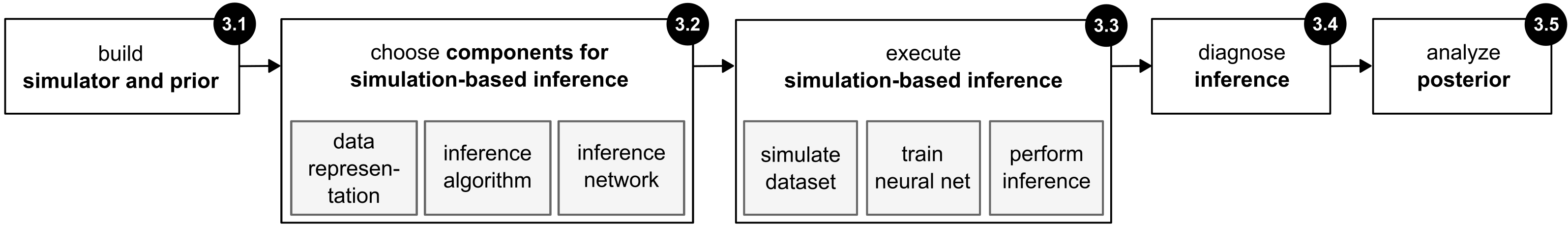}
    \caption{\textbf{Overview of the SBI workflow.} (1) The first step in the workflow is to define the inference problem, which includes setting up the simulator and the prior. (2) Next, we need to choose appropriate components for performing SBI: A representation of the data (raw data, manually designed summary statistics, or embedding networks), an inference algorithm (e.g., NPE, NLE, NRE, ...) and associated inference network (Gaussian, normalizing flow, diffusion model,...).
    (3) Next, we run SBI: We generate the simulated training dataset, train the inference network, and, after training, evaluate the inference network to infer the posterior given an observation. (4) Having obtained the posterior, we validate the inference results with diagnostic tools. (5) If these checks pass, we can analyze the posterior and simulator to gain scientific insight. The numbers at the top right of every box refer to sections in this paper.}
    \label{fig:3_workflow}
\end{figure}

Applying simulation-based inference to a scientific problem involves a structured, sequential workflow, designed to leverage machine learning for robust parameter inference even with complex simulators (Fig.~\ref{fig:3_workflow}).
This process, as outlined in our previous example, typically begins by defining the simulator and the prior over parameters.
Following this, we select the most suitable SBI algorithm and inference network for our data.
The core of the workflow involves generating a large dataset of simulations, training the chosen inference model on this data, and then using the trained model to perform rapid inference on the observed data.
We then undertake validation steps to ensure the reliability and accuracy of the inferred results and we analyze the posterior.
This section elaborates on each of these stages, highlighting key considerations, potential challenges, and practical diagnostic tools to guide practitioners toward successful application of SBI.

\subsection{Defining the problem: Simulator and prior}
\label{sec:3_1_simulator_and_prior}

\paragraph{The simulator: Setting up the forward model}

The \emph{simulator} is the core component in simulation-based inference, and defines the link between the parameters $\btheta$ and the observed data $\bx$.
For inference results to be interpretable and scientifically meaningful, the simulations should be similar to the observed data. If there are no parameter settings which achieve this (a situation referred to as ``misspecification'' \citep{frazier2020model,cannon2022investigating}), this can have a severe impact on inference results. We will return to this question below, and discuss diagnostics and remedies in Sec.~\ref{sec:3_4_diagnostics}.

Importantly, SBI is applicable to any simulator that can be run in the forward direction (i.e., in statistical terms, that can generate samples from the likelihood given parameters: $\bx \sim p(\bx | \btheta)$).
In particular, it can also be applied to ``black box'' simulators, that is, it does not require any access to the inner workings of the simulator, and can therefore also be applied to simulators that do not allow for explicit evaluation of the likelihood or the gradient of the simulation output with respect to the input parameters. This flexibility of SBI is a great advantage with respect to classical Bayesian approaches, which often require making compromises between the scientific realism and the properties of the model for inference to be tractable (e.g., require using a linear model or a Gaussian noise model even if not justified by the properties of the system.)
Moreover, since the inference networks are trained solely on pairs of parameters and simulation outputs, the simulation process and the subsequent neural network training are fully decoupled.
Indeed, simulations can be run offline \emph{before} training begins.
This independence allows simulations to be run in parallel on high-performance computing resources, across different programming languages or computing environments.

\paragraph{Building the prior: Incorporating prior beliefs about parameters}

The \emph{prior distribution} $p(\btheta)$ is another essential ingredient in Bayesian inference \citep{gelman1995bayesian}: 
It describes the prior beliefs, or knowledge, about the parameters for the inference process, before observing any new data. In the standard SBI workflow, we generally sample parameter values $\btheta$ from this prior distribution to generate the training dataset $(\btheta, \bx)$\footnote{However, most SBI methods including NPE can be adapted to sample parameters from so-called ``proposal'' priors which are different from the prior, see Appendix~Sec.~\ref{sec:A2_sequential_methods}.}.

For Bayesian inference, the prior should reflect our prior beliefs, or ideally knowledge, about the  distribution of parameters of the system under investigation. 
For instance, in our ball throwing example, the prior should represent the distribution of plausible throwing angles. This could, for example, be informed by a histogram of throwing angles across many people and throws, or it could even incorporate additional information about the particular person that is throwing.

In scientific problems, it can be informed by prior studies, constraints from physical laws, or expert knowledge. Defining an appropriate prior distribution for a particular scientific problem can be challenging and subtle (for detailed discussions, see, e.g., \citet{winkler1967assessment, confavreux2023meta}).
Many SBI applications use pragmatic ad-hoc choices that reflect basic knowledge about plausible parameters: 
For instance, what is the range in which a parameter is chosen, or is there some known dependency between certain parameters?
A common choice is to select uniform distributions with specified bounds.
The rationale behind these choices is their ease of construction and their perceived neutrality. Nevertheless, it is important to acknowledge that a uniform prior is not truly ``uninformed'' because it implicitly defines an informative choice of scale \cite{bishop_learning_2025}. 

Finally, it is worth noting that one advantage of SBI is its flexibility with respect to the choice of prior: While some classical Bayesian approaches require, or incentivize, use of a particular prior family (e.g., conjugate priors for generalized linear models), for many SBI methods (e.g., NPE) we only need to be able to \emph{sample} from the prior, allowing us to define the prior with focus on its scientific interpretation (rather than tractability of inference). However, and discussed below, the choice of prior might affect how many simulations are required to accurately train inference networks. 

\paragraph{Key considerations}

\begin{wrapfigure}{r}{0.52\textwidth}
  \begin{floatrow}
    \ffigbox[\textwidth]%
      {%
        \raggedleft
        \includegraphics{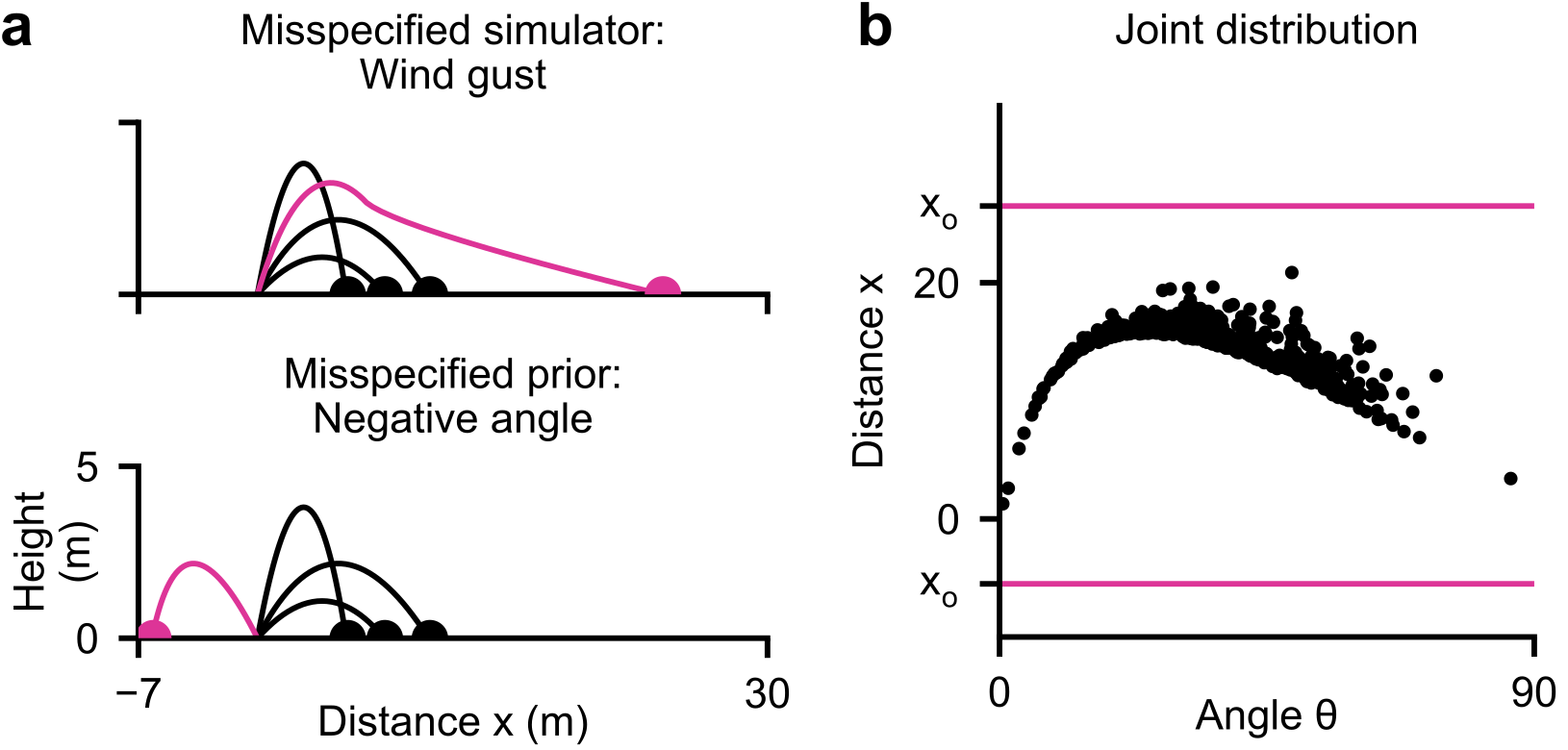}
      }%
      {%
        \caption{%
            \textbf{Model misspecification. (a)} Two examples of misspecified observations (pink) falling outside the range of simulated data (black). Top: The observed distance exceeds all simulations, indicating the simulator does not fully capture real-world conditions---here exemplified by a strong wind gust that produced a larger throwing range. Bottom: The observed distance falls short of all simulations because the prior distribution excludes negative angles, preventing the simulator from generating throws at the actual angle used.
            \textbf{(b)} The joint distribution of simulated data (black) with the two misspecified observations from panel (a) overlaid (pink). Both observations lie outside of the range of the simulated joint distribution.
        }%
        \label{fig:4_misspecification_examples}
      }
  \end{floatrow}
\end{wrapfigure}

When building the simulator and prior, a critical pitfall for the validity of the inference is \emph{model misspecification}, which encompasses both simulator and prior misspecification \citep{cannon2022investigating, schmitt2023detecting, schmitt2024detecting, ward2022robust, kelly2024misspecification, huang2023learning, wehenkel2024addressing, gao2024generalized}. This occurs when the observed data $\bx_o$ falls outside the range of what the model (simulator and prior) can generate.
For example, our ball throwing simulator assumes a relatively weak tailwind.
Thus, a large observed distance generated by a throw with a wind gust causes the simulator to be misspecified with respect to this observation: There is \emph{no} sample from the prior which can generate such a distance (Fig.~\ref{fig:4_misspecification_examples}a, top). Similarly, we restricted the prior to positive angles, but if the observation corresponds to a negative distance (throwing behind oneself, i.e., throwing with a negative angle, Fig.~\ref{fig:4_misspecification_examples}a, bottom), then the prior is misspecified.
In such scenarios, one might expect that SBI produces the posterior distribution given the closest possible data (e.g., the highest simulated distance), or that it simply returns a very uncertain posterior, but this is not the case.
In fact, neural SBI methods can behave erratically and produce unreliable inference results in this setting, often yielding spurious posterior distributions that do not accurately reflect underlying properties of the simulator or the true parameter values \citep{cannon2022investigating}.
Indeed, in the ball-throw cases, the ``cut'' in the observed distribution is empty, and thus wrongly suggests that there is no parameter setting where the model can reproduce the data. (Fig.~\ref{fig:4_misspecification_examples}b).

This consideration is also relevant when selecting the width of the prior. A prior that is too narrow can lead to model misspecification, while a prior that is too broad can significantly increase the computational cost of inference:
Many simulations might be wasted on regions in parameter space that yield results drastically different from any observed data.
Consequently, a broad prior demands substantially more simulations to obtain good results, or, with a fixed simulation budget, it can lead to less accurate inference. Therefore, the width of the prior must be chosen with care to yield valid results while maintaining reasonable computational cost.

To identify model misspecification, a simple a priori check, the so-called \textit{prior predictive check} can be used: One first samples parameters from the prior, runs the simulator for each parameter, and then visually or statistically compares the resulting simulated data to the observed data to verify that the simulated data adequately covers the range of observed data. The data generated for these prior predictive checks can often be reused for training the inference network.
We will describe more elaborate tests for misspecification in Section~\ref{sec:3_4_diagnostics}.


\subsection{Choosing the SBI components: Data representation, inference algorithm and inference network}
\label{sec:3_2_choosing_components}

After defining the simulator and prior, the next step in the SBI workflow is to choose the appropriate components for the inference method. This involves decisions about how to represent the data, which specific SBI algorithm to employ, and what kind of inference network to choose. This section provides a high-level overview of these choices, with a detailed comparison of SBI methods presented in Appendix~Sec.~\ref{sec:A1_choosing_a_method}.

\paragraph{Choosing a data representation: Summary statistics and embedding networks}

Simulators often produce high-dimensional outputs $\bx \in \mathbb{R}^d$ ($d \gg 1$), such as large telescope images in astrophysics or long time series in neuroscience.
In such cases, it is useful---and sometimes essential---to reduce the dimensionality of these observations from $d$ to a lower dimension $r$ (with $r \ll d$) using \emph{summary statistics} $\mathcal{S}: \mathbb{R}^d \to \mathbb{R}^r$.
This dimensionality reduction can serve several purposes: Isolating informative components while discarding noise, eliminating redundant features, and reducing the computational cost of inference. A further motivation for using summary statistics has been to combat model misspecification: In many settings, scientists know that their model can only capture \emph{some} properties of their data which are of particular scientific interest, and they might be aware that there are other properties of their data that the model cannot capture. In such cases, practitioners have traditionally relied on handcrafted summary statistics informed by domain knowledge, to design summary statistics which extract those features of the data which are of both scientific interest and expected to be well-captured by the model~\cite{deistler2022energy,rodrigues2020learningsummaryfeaturestime}.

In general, however, a summary statistic $\mathcal{S}$ should retain as much information about the parameters $\btheta$ as possible, so that the posterior derived from the summary statistics approximates the true posterior: $p(\btheta \mid \mathcal{S}(\bx)) \approx p(\btheta \mid \bx)$. For some simple simulators, optimal summary statistics which preserve all relevant information (known as sufficient statistics) are analytically derivable---for example, the autocorrelation function for Gaussian auto-regressive moving-average models~\cite{rodrigues2020learningsummaryfeaturestime}. However, for complex, nonlinear simulators (e.g., the Hodgkin-Huxley model in neuroscience~\cite{Hodgkin1952} or the Lotka-Volterra system in ecology~\cite{Lotka1910}), sufficient statistics are generally unknown.
When domain expertise is limited or relationships are highly complex, hybrid approaches have been developed that learn informative summary statistics from simulated data~\cite{jiang2017learning,chen2021neural,fearnhead2011constructing}. However, these methods still require a two-stage process: First learning summaries, then performing inference.

For many SBI approaches, one can directly learn summary statistics from data, leveraging the ability of modern deep neural networks to learn representations from high-dimensional data. For NPE, for example, the inference network is a conditional generative model which takes (simulated) data as input and predicts the corresponding posterior distribution. Before passing (simulated) data to this generative model, the data can be passed through a feedforward neural network. This feedforward network can be trained end-to-end with the inference network (e.g., using the log-likelihood loss for NPE with normalizing flows, Eq.~\ref{eq:npe_loss}), and it automatically learns a suitable representation of the data (i.e., summary statistics). Such feedforward networks are known as \emph{embedding networks}.
It can be helpful to think of the entire network to be composed of an \emph{embedding network} $\mathcal{S}_\lambda$ which learns useful representations, as well as the \emph{inference network} $q_{\phi}$ which uses them to perform inference. 

The choice of an appropriate architecture for the embedding network $\mathcal{S}_\lambda$ is a classical machine learning problem, largely depending on the structure and symmetries of the simulated data and observations. For instance, Convolutional Neural Networks (CNNs) are widely used for image data, while Recurrent Neural Networks (RNNs) are effective for time series data. 
For data lacking obvious structure, Multilayer Perceptrons (MLPs) and Residual Networks (ResNets) provide solid baselines. Recently, appropriately designed transformer-architectures have been shown to be highly effective for various types of both structured and unstructured data~\cite{gloeckler2024allinone}. 

An important special case arises when we want to perform inference given not only a single datum $\bx$ as input, but rather a whole data-set of (conditionally) independent observations, and possibly want to even generalize across data-sets of different size: In this setting, Bayesian Inference is ``exchangeable'': The inference result is invariant to permutations of the data. In SBI, such permutation invariance can be achieved using, for example, set transformers as embedding networks~\cite{chan2018exchangeable, lee_set_2019, zaheerDeepSets2018, radev2020bayesflow, von2022mental}. 

Generally, training large-scale embedding networks can require large training datasets and, therefore, many simulations.
For example, while attention-based architectures like Transformers are powerful, their larger number of parameters can sometimes make them impractical if the simulation budget is limited, as they may require a prohibitively large number of simulations to train effectively.

\paragraph{Selecting the inference algorithm}

The core of modern SBI methods lies in training neural networks to perform Bayesian inference. Different algorithms can be categorized based on the quantity that is approximated by the neural network.

A common choice for SBI is \emph{Neural Posterior Estimation} \cite[NPE,][]{papamakarios2016fast, lueckmann2017flexible, greenberg2019automatic, radev2020bayesflow}. As inference network, NPE trains a conditional generative model $q_{\phi}(\btheta | \bx)$ that directly maps each observation $\bx$ to the corresponding posterior distribution $q_{\phi}(\btheta | \bx) \approx p(\btheta|\bx)$.
In principle, NPE can use any conditional generative model: In the simplest case, it is just a multi-layer perceptron (MLP) that takes data as input and learns to predict the means and (co)-variances of a Gaussian distribution over simulator parameters. This can be generalized to Mixtures of Gaussians \cite{Bishop_94, papamakarios2016fast} or to modern, flexible generative models such as normalizing flows \cite{greenberg2019automatic, radev2020bayesflow}, generative adversarial networks \cite{ramesh2022gatsbi}, consistency models \citep{schmitt2024consistency}, flow-matching \cite{lipman2023flow,dax2023flow} or diffusion models\footnote{In the literature, NPE is sometimes considered to only refer to neural posterior estimation with normalizing flows. Here, we use NPE as an umbrella term for all methods that use a generative model to directly estimate the posterior.} \cite{song2021scorebased,geffner2023compositional,sharrock2022sequential}.

A core strength of NPE lies in fully amortized inference: Once trained, NPE can rapidly return samples from the posterior distribution for any new observed data $\bx_o$ without any additional iterative and expensive computations such as MCMC sampling or variational inference which are required by many other methods. 
This property makes NPE highly efficient for scenarios where we anticipate making many inferences on different observed datasets, or for time-critical applications. 
As neural networks excel at learning from high-dimensional inputs, NPE (in combination with appropriately chosen embedding nets) is particularly well suited for inferring parameters from high-dimensional observations $\bx_o$. 
Finally, for problems in which both the input and output dimensionalities are small and many simulations can be generated, NPE often also constitutes an easy-to-use and commonly used approach. 

\emph{Neural Likelihood Estimation} \cite[NLE,][]{papamakarios2019sequential,lueckmann2019likelihood} trains an inference network that approximates the \emph{likelihood} $p(\bx | \btheta)$. This is done with a conditional generative model which also allows for density evaluation (often called a density estimator; for example, a normalizing flow).
Once NLE has learned an approximation of the likelihood, inference for a new observation is performed using classical Bayesian inference methods such as MCMC or VI.
Running MCMC or VI is based on the trained inference network and does not require further simulations.

A key advantage of NLE emerges when dealing with multiple independent and identically distributed (i.i.d.) observations, such as independent trials within a single experiment.
In such cases, the total likelihood for a set of observations can be efficiently computed by multiplying the likelihoods of individual observations
\begin{equation*}
    p(\btheta \mid \bx_o^{1,\ldots,N}) 
    \propto p(\bx_o^{1,\ldots,N} \mid \btheta) \, p(\btheta) 
    = \prod_{i=1}^N p(\bx_o^i \mid \btheta) \, p(\btheta) 
    \approx \prod_{i=1}^N q_{\phi}(\bx_o^i \mid \btheta) \, p(\btheta).
\end{equation*}
Notably, due to this product decomposition, NLE can train on single simulated observations $q_{\phi}(\bx^i \mid \btheta)$ yet perform inference on any number of i.i.d.~observations.
This constitutes an advantage over NPE: NPE must train on $N$ (simulated) observations to perform inference on $N$ observations---typically requiring $N$ simulations per parameter set $\btheta$, making NPE more simulation-hungry than NLE in this scenario.

Another set of methods, \emph{Neural Ratio Estimation} \cite[NRE,][]{cranmer2015approximating,hermans2020likelihood,durkan2020contrastive,miller2022contrastive,thomas2022likelihood}, aims to learn the likelihood ratio $p(\bx|\btheta)/p(\bx)$, which enables using MCMC or VI to obtain posterior approximations. 
The ratio is learned by solving a classification problem---for a pair $(\btheta, \bx)$, a neural network classifies whether $\bx$ was generated by the simulator with parameter $\btheta$ or by a randomly chosen parameter set.
Thus, in NRE, the inference network is a classifier that learns to distinguish between samples from the joint distribution $p(\bx, \btheta)$ and the product of marginals $p(\bx)p(\btheta)$.
The learning task is therefore reduced to a classification problem, which can be computationally cheaper and easier to train than generative models used by NPE and NLE.
In addition, NRE shares the property of NLE to efficiently handle multiple i.i.d.~observations.
However, similar to NLE, obtaining samples from the posterior distribution using NRE requires additional inference methods such as MCMC or VI.

Each of these core methods, NPE, NLE, and NRE, exist in numerous variants. For example, sequential versions iteratively refine the inference to focus on a specific observation $\bx_o$, improving both accuracy and simulation efficiency (see Appendix~Sec.~\ref{sec:A2_sequential_methods}).

Finally, several algorithms also go beyond estimating either the likelihood(-ratio) or the posterior. For example, some methodologies estimate both the likelihood(-ratio) and the posterior using separate neural networks, aiming to leverage the complementary strengths of each approach \citep{wiqvist2021sequential,gloeckler2022variational,radev2023jana}. Recently, the Simformer, a diffusion model based on transformers, has been proposed to estimate the full joint distribution of parameters and data, together with all conditional distributions, including the likelihood and posterior \citep{gloeckler2024allinone}. This allows the Simformer to deal with missing data or variable numbers of i.i.d.~datapoints, and it enables emulation and posterior inference in a single inference network. For more recent trends in SBI, see Appendix Sec.~\ref{sec:A3_recent_developments}.

\paragraph{Choosing the inference network: Normalizing flows, diffusion models, and more}

Beyond the choice of algorithm, selecting an appropriate \emph{inference network} can have an impact on the performance of SBI.
For NPE, the inference network is a conditional generative model that takes as input the data (either raw, summary statistics, or the outputs of an embedding network) and approximates the posterior distribution $p(\btheta|\bx)$.
A popular choice for this inference network are normalizing flows \cite{papamakarios2021normalizing}.
These are powerful architectures that transform a simple base distribution (typically a Gaussian) into a complex target distribution through a series of invertible transformations. These transformations are learned, such that the target distribution approximates the posterior, even if the posterior is non-Gaussian or has higher-order parameter dependencies. After training, normalizing flows can sample from the approximate posterior by sampling from the base distribution and passing the sample through the series of transformations. Normalizing flows can also evaluate the density of a particular parameter set under the posterior using the probability change of variables formula.
Normalizing flows are widely used in NPE for their flexibility in modeling non-Gaussian and multi-modal posteriors, and have often been reported to be easy-to-use in practice, in particular on typical SBI problems with only a handful of parameters. 

More recently, diffusion models and flow-matching have emerged as powerful generative models~\cite{lipman2023flow,song2021scorebased,dax2023flow,geffner2023compositional}.
For SBI, these models offer high flexibility in approximating complex and high-dimensional posterior distributions, but it is typically more computationally expensive to draw samples from the posterior and to evaluate its density.
Beyond these, other generative models have also been explored for NPE---such as tabular foundation models \citep{vetter2025effortless}, consistency models \citep{schmitt2024consistency}, generative adversarial networks (GANs) \citep{ramesh2022gatsbi}, or energy-based models \citep{glaser2022maximum}.
These can perform well on specific tasks or simulation budgets.

In NLE, the inference network takes as input parameters and estimates the likelihood of data $p(\bx|\btheta)$. Similar to NPE, this also involves training a conditional generative model.
A key practical constraint for NLE is the need for efficient evaluation of the log-likelihood in order to perform subsequent inference with MCMC or VI.
This requirement limits the choice of generative models, as sampling methods cannot be used directly.
As such, normalizing flows are a particularly popular choice for the conditional generative model in NLE, as their invertible nature allows for direct and computationally efficient log-likelihood evaluation without further approximations.
Recent research has also explored the use of diffusion models for NLE, leveraging their strong generative capabilities, though evaluating the likelihood with diffusion models is significantly more computationally expensive and can introduce errors~\cite{song2021scorebased,geffner2023compositional,sharrock2022sequential,linhart_diffusion_2024c}.

Finally, for NRE, the inference network is a classifier that is used to obtain likelihood-to-evidence ratios. As such, any classification neural network can be used for NRE. Typically, residual neural networks (ResNets) provide a strong baseline, but specific tasks might require the design of more specific architectures.

\paragraph{Key considerations}

The choice of data representation, inference method, and inference network impact the accuracy and efficiency of SBI. While the array of choices might seem daunting, modern SBI toolboxes~\cite{tejerocantero2020sbi} implement these components in a modular fashion, allowing users to easily switch between different embedding networks, inference networks, and inference algorithms. This modularity enables systematic experimentation---for instance, comparing normalizing flows against diffusion models for NPE, or testing different architectures for the embedding network---all while reusing the same simulated dataset.

Since running simulations is often the most expensive part of the SBI pipeline, simulated data can be reused across different training runs, methods, and architectural choices. This reusability, combined with the plug-and-play nature of modern SBI implementations, allows us to empirically determine the best combination of components for their specific problem without incurring additional simulation costs.

\subsection{Executing SBI: Simulation, training, and inference}
\label{sec:3_3_executing_sbi}

Having selected the SBI components, the next step is to execute SBI.
We first need to generate the simulated training data and then use this data to train the inference network. After training, the inference network is used either to sample parameters from the posterior distribution (in NPE) or as a surrogate in a sampling algorithm to access the posterior distribution (in NLE or NRE).

\paragraph{Generating simulations}

For all SBI algorithms, the inference network is trained on a set of parameter-data pairs $(\btheta,\bx)$.
This dataset is generated by first sampling a set of parameters $\btheta$ from the prior distribution (see Sec.~\ref{sec:3_1_simulator_and_prior}), and the simulator is then run with these parameters to generate a corresponding synthetic observation $\bx$.
If the simulator is expensive to run, only a limited number of simulations might be feasible. As discussed below, an insufficient number of simulations can lead to poor inference results. 
Notably, generating simulations is decoupled from the training process, which enables an offline approach for running simulations: One can generate a large database of simulations once and then reuse it to train different neural networks, architectures, or inference methods without re-running the computationally expensive simulations. In addition, simulations can be generated in parallel, leveraging batching or other techniques to accelerate large-scale data generation~\cite{Meyer2023}.

\paragraph{Training the neural inference model}

\begin{wrapfigure}{r}{0.59\textwidth}
  \vspace{-0.5cm}
  \begin{floatrow}
    \ffigbox[\textwidth]%
      {%
        \raggedleft
        \includegraphics{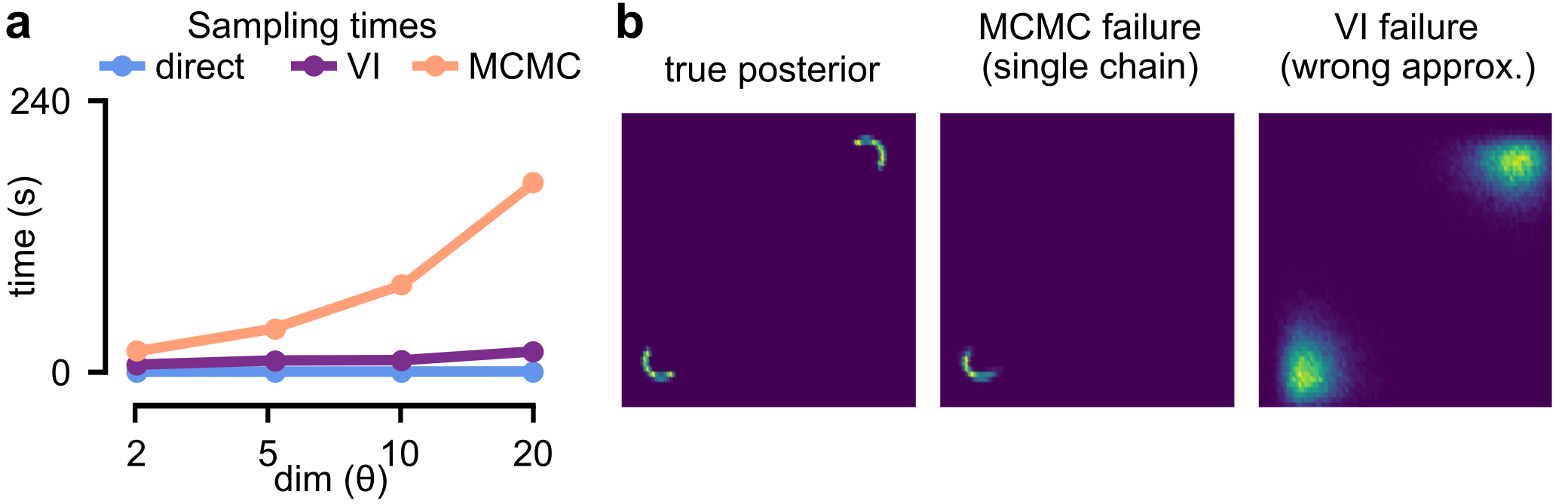}
      }%
      {%
        \caption{%
        \textbf{Example demonstrating computational runtime and accuracy of sampling methods.}
          \textbf{(a)} Runtimes to acquire 1000 posterior samples in a linear Gaussian model for NPE with normalizing flows (here called ``direct''), and for NLE with variational inference (VI) and Markov chain Monte Carlo (MCMC). Evaluated on a Core i7-11800H CPU.
          \textbf{(b)} Potential failures of MCMC or VI.
          Left: True posterior distribution.
          Middle: NLE samples with poorly tuned MCMC (e.g., using only a single chain). Although the inference network trained by NLE has converged well (not shown), MCMC sampling fails.
          Right: NLE samples with poorly tuned VI (e.g., poor convergence). Like MCMC, VI can also introduce errors to the sampling procedure required by NLE or NRE.
        }%
        \label{fig:5_mcmc_or_vi_failures}
      }
  \end{floatrow}
\end{wrapfigure}

Proper training of the neural network (including inference network and, if used, embedding network) is essential for reliable SBI results.
A fundamental consideration is appropriate splitting of simulated data into training, validation, and test sets, mirroring best practices in general machine learning applications.
During the training phase, monitoring the training and validation loss curves is crucial to identify convergence and avoid overfitting.
It is therefore advised to rely on regular checkpointing and early stopping during training to reduce the chance of unsuccessful and wasteful training.
Many software packages (e.g., \texttt{sbi} \cite{boelts2024sbi}, sbijax \cite{dirmeier2024simulation}, Bayesflow \cite{radev2023bayesflow}, or swyft \cite{miller2022swyft}) implement pre-configured training loops that effectively manage such tasks.
However, for achieving optimal performance and fine-tuning the neural network architecture or chosen generative model, further optimization of hyperparameters (such as batch size, learning rate, or the early stopping schedule) is often desirable.
We recommend using experiment tracking (e.g., MLFlow \cite{zaharia2018accelerating}, WandB \cite{wandb}, Hydra \cite{yadan2019hydra}) tools to systematically compare different architectures, hyperparameter settings, and training runs.

For simulators with a moderate number of parameters (e.g., less than $\approx 100$) and data dimensionality (e.g., less than $\approx 100$), neural network training is typically fast, often lasting on the order of minutes. In such cases, training on a CPU can be as fast as or even faster than on a GPU, particularly for batch sizes below 1000. GPUs primarily offer significant speed-ups for large batch sizes or very high-dimensional data (especially when combined with a suitable embedding network).

\paragraph{Performing amortized inference}

After the inference network has been successfully trained on simulated data, the next step in the SBI workflow is to perform inference on one or multiple observed datapoints $\bx_o$.

A key advantage of most neural SBI methods is that they are \emph{amortized}. This means that once the initial phase of generating simulations and training the inference network is complete, the trained neural network can be applied to any new observation without requiring further simulations or retraining. For NPE in particular, obtaining posterior samples is exceptionally fast, often generated in milliseconds through a single forward pass of the network. 

For other neural SBI methods that do not directly output the posterior distribution (e.g., NLE or NRE), performing inference on a new observation requires an additional step, often involving standard Bayesian inference methods like Markov chain Monte Carlo (MCMC) sampling or Variational Inference (VI).
These methods use the trained inference network to efficiently evaluate the approximate likelihood or likelihood-ratio as required by MCMC or VI.
While these sampling steps typically do not incur the high computational cost of running the simulator, they can still take several seconds or minutes, especially for models with many parameters (Fig.~\ref{fig:5_mcmc_or_vi_failures}a).
Moreover, these sampling algorithms come with their own sets of hyperparameters that need careful tuning for the specific task.
If not set correctly, they can lead to performance degradation or incorrect results, independent of the quality of the trained inference network (Fig.~\ref{fig:5_mcmc_or_vi_failures}b).
Examples of such issues include MCMC chains failing to converge or VI methods collapsing to a single mode.
These pitfalls highlight the importance of properly configuring and evaluating the underlying sampling algorithms.
Dedicated tutorials exist on MCMC and VI, for detailed guidance on assessing convergence, mixing, and other evaluation metrics (e.g., visual trace plots, effective sample size, R-hat convergence diagnostic) \citep{gilks1995markov, blei2017variational}.

\paragraph{Key considerations}

\begin{wrapfigure}{r}{0.35\textwidth}
  \vspace{-0.5cm}
  \begin{floatrow}
    \ffigbox[\textwidth]
      {
        \raggedleft
        \includegraphics{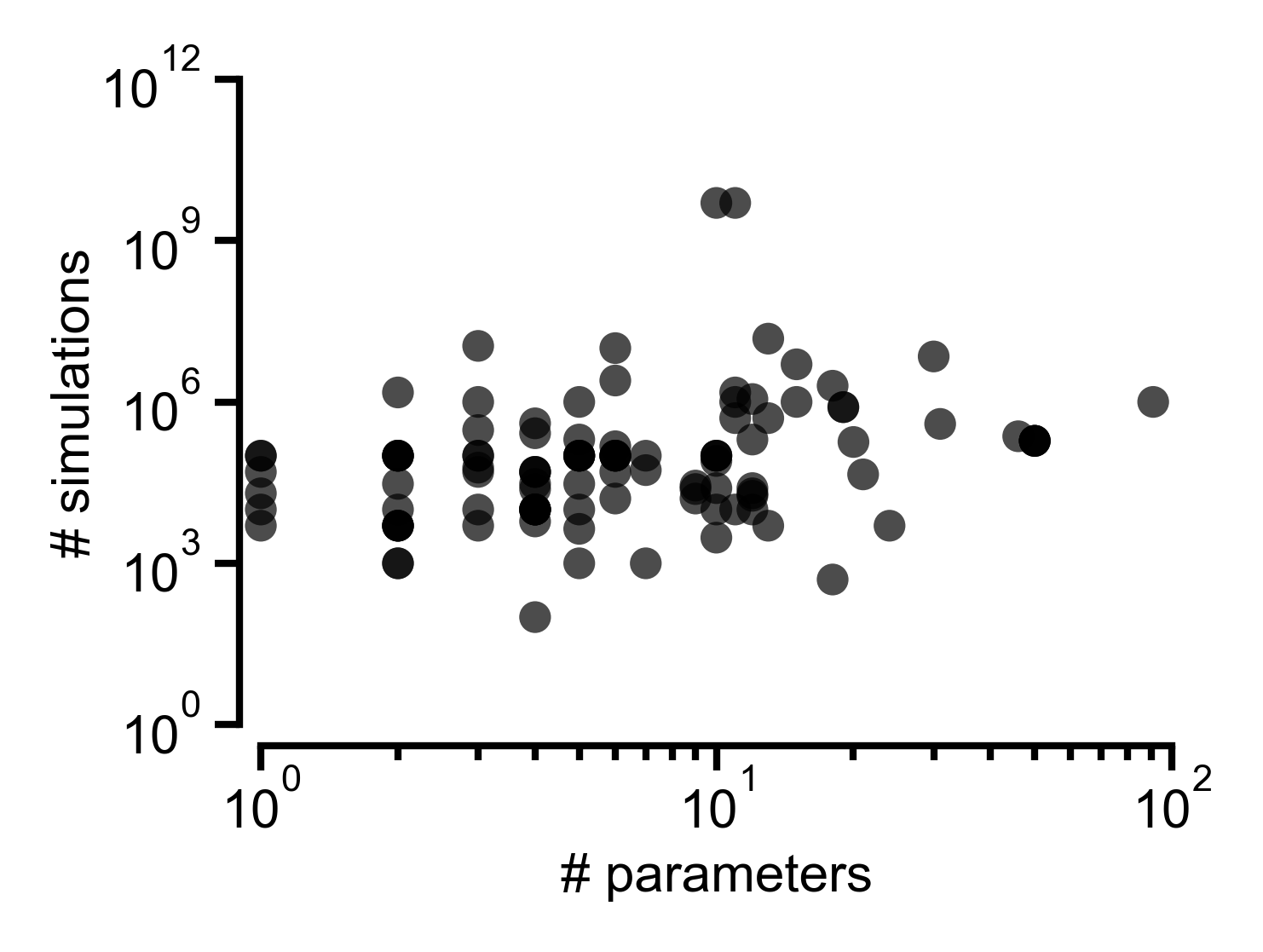}
      }
      {%
        \caption{%
            \textbf{Published SBI applications.} Number of simulator parameters vs number of simulations used for a set of 106 published SBI applications. Each point represents a distinct inference task. 
            An interactive version is available at \href{https://sbi-applications-explorer.streamlit.app/}{sbi-applications-explorer.streamlit.app}.
        }
        \label{fig:6_sbi_applications}
      }
  \end{floatrow}%
\end{wrapfigure}

A key question for SBI is the number of simulations that are necessary to train the model.
While most successful applications of simulation-based inference have required at least a few thousand simulations, there is no single answer to the number of required simulations, as it depends on factors such as simulator complexity, data dimensionality, and architecture size.
In addition, a larger number of parameters increases the search space exponentially, which can also lead to a drastic increase in the number of required simulations.
To illustrate the difficulty of providing guidance on the number of required simulations, we have compiled a database of published SBI applications across diverse scientific domains\footnote{See \url{https://sbi-applications-explorer.streamlit.app/} for an interactive version that allows exploring the results. The web application also provides detailed metadata for each application and enables community contributions to keep the database current. Implementation details and data collection methodology are described in Appendix~Sec.~\ref{sec:A5_sbi_database}.}.
Within these published papers, there is no significant correlation between the number of parameters and the number of simulations (Fig.~\ref{fig:6_sbi_applications}).
This highlights how factors beyond parameter count critically influence computational requirements.

Beyond generating enough data to enable effective network learning, it is equally important to ensure that the validation and test sets are sufficiently large and representative to robustly evaluate the trained model, as inadequate sizes can lead to unreliable assessments of its generalization capabilities.

\subsection{Diagnostics for the posterior distribution}
\label{sec:3_4_diagnostics}

\begin{figure}[t!]
    \centering
    \includegraphics[width=\textwidth]{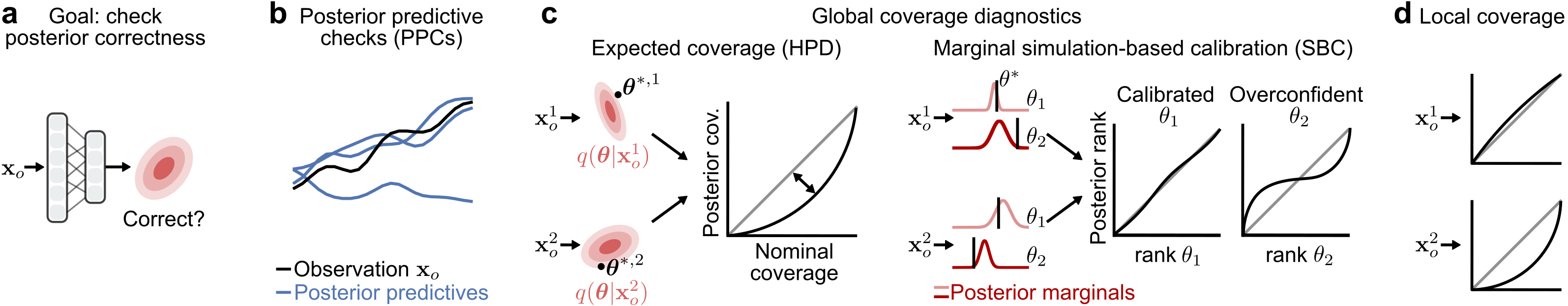}
    \caption{\textbf{Posterior diagnostics.}
    \textbf{(a)} The goal of posterior diagnostics is to validate if an estimated posterior $q_{\phi}(\btheta|\bx_o)$ from a trained SBI inference network is correct.
    \textbf{(b)} Posterior predictive checks (PPCs) compare simulation results of posterior samples with the observed data $\bx_o$.
    \textbf{(c)} Global coverage diagnostics. Left: Given a test set of parameters $\btheta^{*,i}$ and simulated data $\bx^i_o$, global expected coverage assesses whether the parameters fall within the confidence regions of the posterior at the expected rate. Right: Marginal Simulation-Based Calibration (SBC) assesses coverage through ranking within each 1D marginal dimension of the posterior, yielding rank plots for each parameter dimension.
    \textbf{(d)} Local coverage checks assess the coverage for specific observed data.
    }
\label{fig:7_diagnostics}
\end{figure}

Following the previous steps, we have obtained an estimated posterior distribution for an observation $\bx_o$. But how can we judge whether this approximation is accurate (Fig.~\ref{fig:7_diagnostics}a)? Assessing the quality of the posterior estimator is necessary for any downstream analysis and interpretation.

A fundamental limitation of comprehensive diagnostic tools is that, for almost any real-world simulator, the ground-truth posterior distribution is unavailable.
Therefore, we cannot simply compare the posterior estimator obtained via SBI directly to the true posterior distribution. To overcome this, several strategies for diagnosing posterior estimates based solely on the information that is available have been developed. All these methods share a drawback: They require additional simulations beyond the initial training set to compose what is usually called a ``calibration set''. As such, performing them can consume a significant part of the overall computational budget of the SBI workflow.
Below, we outline a range of diagnostic methods: Checks for model misspecification, posterior predictive checks (PPCs), and global and local coverage checks.

\paragraph{Checking for model misspecification}

Model misspecification occurs when the observed data $\bx_o$ cannot be generated by the simulator, even with ideal parameters and accounting for stochasticity (see Sec.~\ref{sec:3_1_simulator_and_prior}). One way to detect this is checking whether $\bx_o$ is within the distribution of simulated data used for training.

The simplest check is a prior predictive assessment: Visualize the training data distribution (e.g., through histograms of summary statistics) and verify that $\bx_o$ falls within the covered ranges~\citep{bernaerts2025combined}. While quick, this approach may miss higher-order discrepancies. A more thorough test involves training an unconditional density estimator to estimate $p(\bx)$ using the simulated data and then evaluating the likelihood of $\bx_o$ under the trained estimator. If this likelihood is exceptionally low compared to all training samples, the simulator is likely misspecified. Crucially, both tests can be performed \emph{before} training any inference network.

For high-dimensional data, more sophisticated approaches include out-of-distribution detection in the latent space of an embedding network~\citep{schmitt2023detecting} or, when likelihood evaluation is possible, computing the evidence $p(\bx_o)$ via importance sampling using the posterior as proposal~\citep{dax2023neural}.

\paragraph{Posterior predictive checks}

Posterior predictive checks (PPCs) provide a rough assessment of the overall fit of the posterior estimator by comparing samples from the posterior predictive distribution to the observation $\bx_o$. This is achieved by drawing $N$ samples from the estimated posterior distribution for a given observation $\bx_o$, and then running the simulator for each of these sampled parameter sets. The resulting ``posterior predictives'' (i.e., the simulation outputs generated from posterior samples) are compared to the original observed data $\bx_o$ (Fig.~\ref{fig:7_diagnostics}b, left). A good posterior estimate should produce posterior predictives that resemble the observed data (taking into account simulator noise). This comparison often includes ``prior predictives'' (simulations from prior samples, e.g., the training data) to assess how much the posterior has updated the data fit compared to the prior alone. While PPCs do not provide sufficient conditions for posterior correctness (e.g., an approximate posterior collapsed on the maximum-likelihood estimate might pass them perfectly), they are useful for quickly detecting gross misfits or qualitatively assessing if the simulator can replicate key features of the observed phenomena. If PPCs fail, it is a strong indicator of issues, pointing to a too small training dataset or to model misspecification.

\paragraph{Coverage diagnostics}

Beyond predictive checks, a more rigorous assessment of posterior quality involves \emph{coverage diagnostics}. These methods check the statistical calibration properties of the estimated posterior, ensuring that its uncertainty estimates are neither overconfident nor underconfident. Coverage diagnostics are categorized into two main approaches: global and local. We briefly outline popular coverage diagnostics below and provide details on these tools in Appendix~Sec.~\ref{sec:A4_diagnostic_tools}.

\emph{Global coverage checks} assess the calibration of the posterior distribution \emph{on average} across many observations (Fig.~\ref{fig:7_diagnostics}c). They are computationally cheaper than local coverage checks and provide a quick overall quality check. However, they might not detect issues if errors average out across observations or if SBI returns an uninformative posterior (in particular, the prior distribution passes these checks).

The most common global diagnostic tools are \emph{Expected Coverage Tests based on Highest Posterior Density (HPD)} \cite{hermans2022crisis} and \emph{Simulation-Based Calibration (SBC)} \cite{talts2018validating}\footnote{Expected coverage tests can even be considered a special case of SBC.}. Both methods typically begin by drawing $N$ parameters from the prior and running the simulator once for each of them, creating a calibration dataset of $N$ pairs of ground truth parameters and simulated data.

Inference is then performed for each data point in this calibration set, yielding $N$ estimated posterior distributions. Each posterior distribution is then checked against the parameter that generated the simulation output. How exactly this check is performed differs between expected coverage and SBC (details in Appendix~Sec.~\ref{sec:A4_diagnostic_tools}), thus yielding different interpretations of the test results.
Expected coverage allows assessing whether the joint posterior is, on average across observations, too narrow or too broad. Because it checks the joint posterior distribution, the diagnostic will indicate issues even if only higher-order moments of the posterior are miscalibrated (e.g., incorrect parameter correlations). However, expected coverage cannot pinpoint which parameters (or higher order moments) caused the detected issues. In contrast, SBC allows assessing whether the posterior \emph{marginals} (i.e., individual parameter dimensions) are too narrow, too broad, or skewed to the left or to the right. Hence, it can be used to assess which parameters have calibration issues. However, the test will pass if all marginals are well-calibrated but higher-order moments are miscalibrated. For HPD and for SBC, the test results are histograms over the test statistics, that can be visually or quantitatively evaluated (Fig.~\ref{fig:7_diagnostics}c).
An alternative global diagnostic, Tests of Accuracy with Random Points (TARP) \cite{lemos2023sampling}, avoids some limitations of HPD-based methods by using randomly positioned credible regions. Through this, TARP can also be combined with methods based on generative models that do not allow for density evaluation.

\emph{Local coverage checks} aim to detect miscalibration for \emph{specific individual observations} (Fig.~\ref{fig:7_diagnostics}d). They are generally more powerful for pinpointing subtle failures that global checks might miss. However, the computational cost of local coverage checks is typically much higher, often requiring significantly more simulations and potentially training additional neural networks for diagnosis. As such, local diagnostics can also introduce their own errors, for example, due to imperfect convergence of the neural network. Recent methods for local coverage checks include Local Coverage Test (LCT) \cite{zhao2021diagnostics}, Local Classifier-two-sample testing (L-C2ST), or posterior SBC \citep{sailynoja2025posterior}.

All coverage diagnostic methods (both global and local) require generating a calibration set comprised of prior samples and corresponding simulations. Subsequently, inference must be performed for every sample in this calibration set. This process critically benefits from \emph{amortized inference} (e.g., via NPE), which allows rapid posterior evaluation for each sample. Conversely, SBI methods that do not directly output the posterior (e.g., NLE or NRE) will incur a significantly larger computational burden for these diagnostic checks, as they require sampling with MCMC or VI for every sample in the calibration set.

\paragraph{Key considerations}

Diagnosing the correctness of the inference result is an important---but sometimes neglected---part of the SBI workflow. While the diagnostic tools presented offer robust ways to assess posterior quality, they also have limitations. Global diagnostics typically provide only necessary conditions for posterior correctness, meaning that passing all global checks does not guarantee that the posterior is accurate. Local diagnostic tools, especially those that rely on training additional neural networks, can themselves be prone to errors, potentially providing misleading diagnostic results.

If any of the above diagnostic methods suggest issues, we should revisit and improve the inference pipeline. This might involve increasing the number of simulations, choosing a different SBI method or inference network architecture, refining the training loop hyperparameters, or, in case of misspecification, changing the prior or even the simulator.

\subsection{Analyzing the posterior}
\label{sec:3_5_using_the_posterior}

Once a robust and well-diagnosed posterior distribution has been obtained, the final and most crucial step in the SBI workflow is to leverage this distribution to gain scientific insight, analyze the underlying model, and inform decision-making. The full posterior provides a rich representation of parameter uncertainty, dependencies, and plausible solutions, enabling a deeper understanding of the complex phenomena modeled by the simulator.

\paragraph{Visualizations}
A first step in understanding the obtained posterior distribution is to visualize it (for examples, see Fig.~\ref{fig:8_grav_wave}a, \ref{fig:9_ddm}c, \ref{fig:10_pyloric}g,h). However, visualizing the full (joint), high-dimensional posterior is often not possible. Therefore, we typically visualize 1D or 2D marginal posterior distributions (i.e., the projection on one/two dimensions), or 1D or 2D conditional posterior distributions (i.e., distributions in which all but one/two parameters are fixed). Such visualizations enable an intuitive inspection of posterior properties and the relationships between parameters (e.g., correlations, compensation mechanisms, or non-identifiabilities).

\paragraph{Studying the posterior parameter space}
Going beyond visualizing the posterior, we can also quantitatively study the parameter space captured by the posterior. For example, by repeatedly performing inference for a subset of summary statistics, we can evaluate which features of the data most constrain parameters \citep{beck2022efficient}, or one can study the sensitivity of the posterior to changes in the likelihood or prior \citep{elsemuller2024sensitivityaware}. Other potential analyses are to evaluate whether the posterior is unimodal or multimodal \citep{gonccalves2020training}, how the posterior distributions vary across observations, or how sensitive simulation outputs are to parameter changes, for instance through posterior predictive checks or local perturbations in parameter space.

\paragraph{Marginal moments: Mean, standard deviation, and covariance}
Another analysis is to inspect moments of the posterior such as its marginal mean, standard deviation, or covariance. As sampling from the posterior is usually cheap, this can be done easily with a Monte--Carlo estimate as
\begin{equation}
    \mathbb{E}_{p(\btheta | \bx_o)}[f(\btheta)] = \int p(\btheta | \bx_o) f(\btheta) \,\text{d}\btheta \approx \frac{1}{N} \sum_i^N f(\btheta_i),
\end{equation}
where $\btheta_i$ are samples from the posterior and $f(\btheta_i)$ is a function of parameters. Analyzing the mean can already provide an estimate of trends in the posterior distribution, the standard deviations enable analyses into how constrained the posterior is, and the covariances allow studying interactions between parameters.

\paragraph{Conditional moments: Mean, standard deviation, and covariance}
Studying marginal standard deviations and covariances can, however, lead to wrong conclusions regarding how strongly tuned individual parameters have to be. For example, due to parameter correlations (or higher-order interactions), high standard deviations might not mean that a parameter has to be entirely untuned. 
To investigate this, in addition to the marginal mean, standard deviation, and covariances, SBI methods can also compute the \emph{conditional} mean, conditional standard deviation, and conditional covariances as
\begin{equation}
    \mathbb{E}_{p(\btheta_{i} | \btheta_{/i}, \bx_o)}[f(\btheta_i)] = \int p(\btheta_{i} | \btheta_{/i}, \bx_o) f(\btheta_i) \,\text{d}\btheta_i,
\end{equation}
where $\btheta_{/i}$ is a parameter vector without the i-th dimension. 
This integral can be solved with quadrature (because the integral is 1D for conditional mean and standard deviation and 2D for the conditional covariance). The conditional mean and variance allow us to study the ranges of parameters \emph{given the values of other parameters}. This can enable insights into how finely parameters have to be co-tuned to match the observation $\bx_o$, or into compensation mechanisms between parameters.

\paragraph{Maximum-a-posteriori (MAP) estimation}
While the full posterior distribution provides a comprehensive view, practitioners often seek a single ``best'' parameter estimate. The \emph{Maximum-a-posteriori (MAP)} estimate is a common choice, representing the mode (the parameter value with the highest probability density) of the posterior distribution. Since (most) SBI methods yield posteriors that can evaluate the (unnormalized) posterior log-probability and are differentiable, finding the MAP estimate can be achieved efficiently through optimization methods (e.g., gradient ascent).

\paragraph{Bayesian decision theory}
Beyond parameter analysis, the full posterior distribution can be used for decision-making under uncertainty. \emph{Bayesian decision theory} provides a formal framework for choosing optimal actions by explicitly accounting for all available knowledge (encoded in the posterior) and the consequences of different choices. Bayesian decision theory chooses an action $a$ which minimizes
\begin{equation}
    \ell(\bx_o, a) = \int p(\btheta | \bx_o) \cdot c(\btheta, a) \d \btheta,
\end{equation}
where $c(\btheta, a)$ is a cost function which defines the cost of performing action $a$ if the true parameter set is $\btheta$. This framework can be applied to diverse problems in science and engineering where decisions need to be made based on uncertain parameter estimates from simulation-based models \cite{berger2013statistical, gorecki2024amortized, alsing2023optimal}.

\section{Examples}
\label{sec:4_examples}

We have outlined a full workflow on how to obtain the posterior distribution with SBI, as well as how to choose between different existing inference methods. We will now apply this workflow to three inference problems: A gravitational wave model from astrophysics to demonstrate the ability of NPE to perform amortized inference and use an embedding network, a drift-diffusion model from psychophysics to demonstrate the ability of NLE to efficiently handle i.i.d.~datapoints, and a biophysical model from neuroscience to demonstrate the ability of SBI to scale to dozens of parameters. These problems illustrate the workflow of SBI and its capabilities.

\subsection{Amortized inference on a simplified gravitational wave model from astrophysics}
\label{sec:4_1_grav_waves}

\begin{figure}[t!]
    \centering
    \includegraphics[]{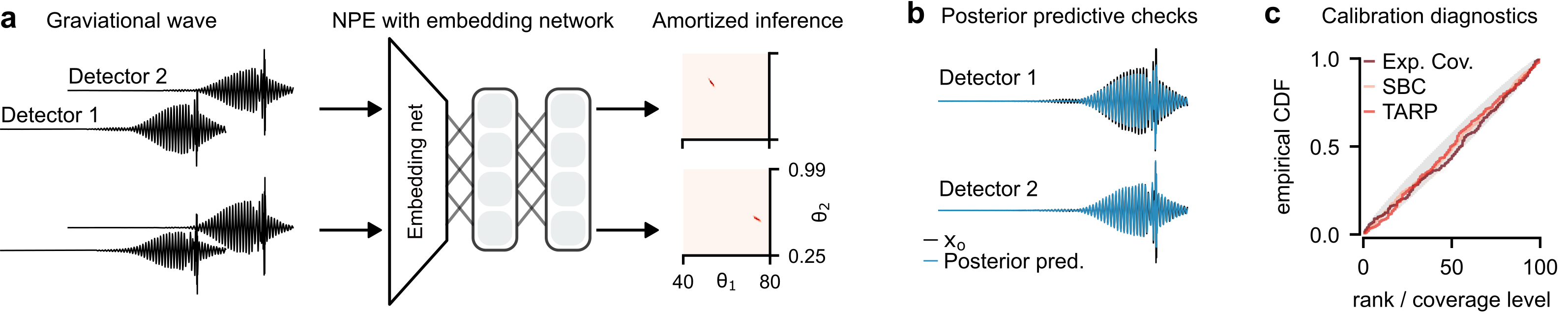}
    \caption{\textbf{Amortized inference for gravitational wave time series.}
    \textbf{(a)} Left: Two synthetic gravitational wave signals of a binary black hole system, each recorded by the two LIGO detectors.
    Middle: NPE with embedding network.
    Right: Posterior distributions for the two parameters (masses of the black holes) given each of the two synthetic gravitational wave signals. For both observations, the posterior is constrained to small regions in parameter space (dark red spots).
    \textbf{(b)} Observation $\bx_o$ and posterior predictive simulations of an NPE posterior sample.
    \textbf{(c)} Calibration checks (expected coverage, TARP, marginal SBC, with confidence regions in gray) demonstrate that the NPE network is well-calibrated.
    }
    \label{fig:8_grav_wave}
\end{figure}

Our first example demonstrates how SBI enables real-time analysis of gravitational wave—ripples in spacetime caused by accelerating masses. When two black holes spiral around each other and eventually merge, they emit gravitational waves that can be detected on earth by observatories like the Laser Interferometer Gravitational-Wave Observatory (LIGO). Here, we use a simplified simulator that models these gravitational wave signals as recorded by two detectors, with the goal of inferring the masses of both black holes (our two parameters) from the observed time series. This scenario showcases two key strengths of SBI: Handling high-dimensional observations (time series with thousands of data points) and providing amortized inference that enables rapid analysis of multiple events, critical for real-time astronomical discovery \citep{dax2021real}. It constitutes a simplified setup relative to full analysis approaches for GW models, which target multiple model parameters and additional data-properties with augmented models and analysis pipelines (see, e.g., \cite{dax2021real,dax2023neural, WildbergerPRD,dax2025real,dax2022group}). 

Our simulator generates two time series of length $8192$ for each observation of a black hole merging event (Fig.~\ref{fig:8_grav_wave}a, left). These time series represent the signal as it would be recorded by the LIGO Hanford Observatory $H1$ and LIGO Livingston Observatory $L1$ \citep{AdvancedLIGO2015}. We assumed flat uniform priors for both parameters within the interval of $[40,80]\,M_{\odot}$ (solar masses). To constrain the second black hole mass, we limited its value to $[0.25,0.99]\cdot\theta_1$ to ensure that it must be always lighter than the first. We generated a total of $75k$ simulations.

As we aimed to perform fast inference for many observations, we used NPE, with a neural spline normalizing flow \cite{durkan2019neural} as inference network. Due to the high dimensionality of the observations, we used  an embedding network for automatically reducing the available information in the time domain (Fig.~\ref{fig:8_grav_wave}a, middle). To achieve this, we used a 1D convolutional neural network as embedding network (details in Appendix Sec.~\ref{sec:A4_1_grav_wave_details}). The embedding network reduces the dimensionality of $2\times8192$ timepoints to a $32$-dimensional latent representation of each observation.

After training, we were able to perform inference for several synthetic observations within milliseconds due to the amortization property of NPE. As we only apply the inference network to synthetic observations (i.e., simulations with parameters from the prior), we skip any check for model misspecification. The resulting posteriors were narrow (compared to the prior), suggesting that the gravitational wave data strongly constrain the masses of black holes given this simulator (Fig.~\ref{fig:8_grav_wave}a, right). We then validated the inference result. We first performed posterior predictive checks and found that simulations of posterior samples closely matched the observation (Fig.~\ref{fig:8_grav_wave}b). Next, we validated the NPE network with several diagnostic metrics. Expected coverage, SBC as well as TARP (Appendix Sec.~\ref{sec:A4_diagnostic_tools}) indicated that the posterior was well-calibrated (Fig.~\ref{fig:8_grav_wave}c).

Overall, this example demonstrates that NPE can be used to perform inference given high-dimensional observations by using an embedding network, and that the ability to perform amortized inference enables real-time analysis of observed phenomena.

\subsection{Inference given i.i.d trials on a drift-diffusion model from psychophysics}
\label{sec:4_2_ddm}

\begin{figure}[t!]
    \centering
    \includegraphics[width=\textwidth]{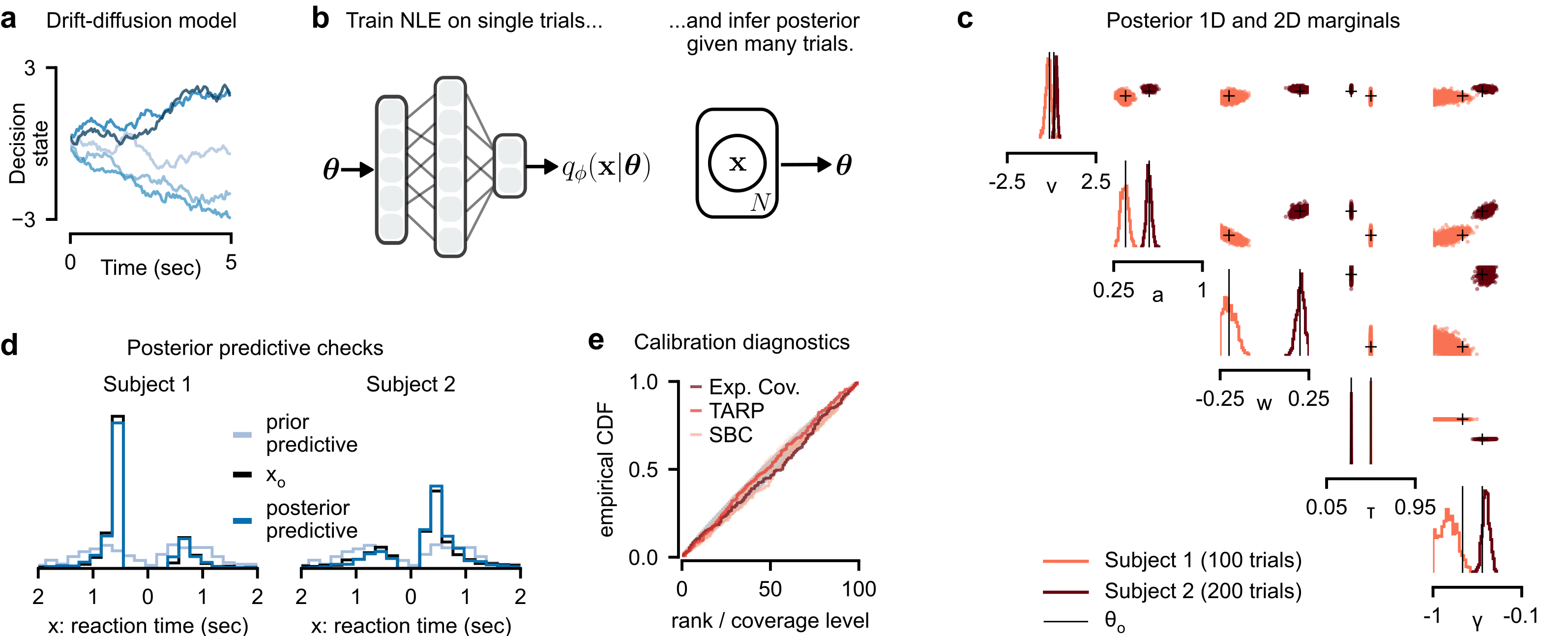}
    \caption{\textbf{Inference on the drift-diffusion model with trial-based i.i.d. data.}
    \textbf{(a)} Prior predictive simulations of the drift-diffusion model. The reaction time (RT) is recorded once the decision state hits a defined boundary. 
    \textbf{(b)} NLE inference networks can be trained on single trials (i.e., they require only a single simulation per parameter set), but can then perform inference for arbitrary numbers of trials.
    \textbf{(c)} The five-dimensional DDM posterior for two subjects (light and dark red), visualized as a corner plot: Diagonal panels show 1D marginal distributions for each parameter; upper off-diagonal panels show 2D marginals for parameter pairs. The tighter distributions for subject 2 reflect reduced uncertainty from observing more trials.
    \textbf{(d)} Posterior predictive checks for both subjects plotted as histogram over RTs. We plot RTs for choice 1 to the left and for choice 2 to the right of 0.  Posterior predictives are closer to the observation $\bx_o$ than prior predictives.
    \textbf{(e)} Calibration checks (expected coverage, TARP, marginal SBC, with confidence regions in gray) demonstrate that the NLE network is well-calibrated.
    }
    \label{fig:9_ddm}
\end{figure}

Our second example features a drift-diffusion model (DDM), commonly used in psychophysics and cognitive neuroscience to simultaneously model choices and reaction times in decision tasks \citep{ratcliff_diffusion_2008,gold_neural_2007}. Intuitively, the DDM assumes that decisions arise from an internal process where noisy evidence favoring one alternative over another is accumulated over time until a threshold is reached, triggering a response (choice) at a specific reaction time (RT) (Fig.~\ref{fig:9_ddm}a). While the likelihood function can be computed analytically for the simplest DDM variants, many extensions incorporating more realistic cognitive mechanisms---such as trial-to-trial variability in parameters, time-varying evidence (dynamic drift rates), or collapsing decision boundaries (where the evidence required decreases over time)---result in models with intractable likelihood functions. Here, we specifically used a DDM variant with collapsing decision boundaries, parameterized by drift rate ($v$), initial boundary separation ($a$), starting point ($w$), non-decision time ($\tau$), and collapse rate ($\gamma$) \cite{hawkins2015revisiting}. This example illustrates how SBI handles a frequent experimental scenario: Analyzing data composed of multiple independent and identically distributed (i.i.d.) trials, where the number of trials might differ across subjects or observations and that contains both discrete choices and continuous RTs as parameters.

For settings with trial-based data, Neural Likelihood Estimation (NLE) or Neural Likelihood-ratio Estimation (NRE) methods are attractive because their neural networks can be trained on simulations of single trials. Subsequently, for inference on a dataset comprising multiple i.i.d.~trials from a specific subject (with potentially varying numbers of trials), the log-likelihood for the entire dataset is efficiently obtained by summing the single-trial log-likelihoods estimated by the pre-trained network (Fig.~\ref{fig:9_ddm}b). This estimated total log-likelihood can then be used within standard MCMC or VI algorithms to obtain the posterior distribution for individual subjects. This approach avoids retraining the network for each subject or dataset size, making it highly efficient for typical experimental workflows.

Following the SBI workflow, we chose a uniform prior distribution for the five DDM parameters with ranges as common in the literature \cite{hawkins2015revisiting}.
For modeling the mixed (e.g. discrete and continuous) data of the DDM, we employed NLE with a specific generative model designed for this setting \citep{boelts2022flexible}. 
After training the network using 200,000 single trials, we performed MCMC inference using the summed log-likelihoods for two simulated subjects with 100 and 200 trials, respectively (see Appendix~\ref{sec:A4_2_ddm_details} for details). The resulting five-dimensional posteriors accurately recovered the ground-truth parameters for both subjects, with uncertainty estimates varying across parameters and decreasing with more trials (Fig.~\ref{fig:9_ddm}c). Following the SBI workflow, we then performed validation checks. Posterior predictive checks showed that for both subjects the observed data were well-captured by simulations from the inferred posteriors (Fig.~\ref{fig:9_ddm}c). Furthermore, calibration diagnostics performed on a separate dataset (200 additional simulations) indicated overall well-calibrated posteriors: SBC indicated uniform ranks of empirical CDFs for all marginals, alongside well-calibrated joint coverage assessed by expected coverage and TARP (Fig.~\ref{fig:9_ddm}e).

Our choice of NLE for this example highlights a common trade-off in SBI. While training NLE is rather simulation-efficient as it only requires single-trial simulations, the inference step necessitates running MCMC for each subject individually and usually is slower (e.g., minutes for 1,000 posterior samples given 200 trials) and less accurate with increasing number of trials. This makes inference and subsequent diagnostics computationally more demanding, especially when analyzing many subjects or performing extensive calibration checks like SBC, which requires running several MCMC inferences in our case. 
If we used NPE as an alternative, we would have the advantage of extremely fast amortized inference and diagnostics, as obtaining posterior samples only requires a single forward pass through the trained network \cite{radev2020bayesflow}. However, training NPE for i.i.d.~data is more involved and computationally costly, as it requires running as many simulations \emph{per parameter set} as the maximum number of trials expected in the observed data. In addition, it also requires specific permutation-invariant network architectures, data preparation, and requires rerunning simulation and training if the experimental setup changes or the data exceeds the maximum trial number. 
Thus, the choice between NLE and NPE often involves balancing the upfront cost and complexity of simulation and training (high for NPE) against the computational expense during inference and validation (high for NLE) (see also Appendix Tab.~\ref{tab:method_comparison}).

\subsection{Inference of dozens of parameters of a neuroscience model}
\label{sec:4_examples:pyloric}

\begin{figure}[t!]
    \centering
    \includegraphics[]{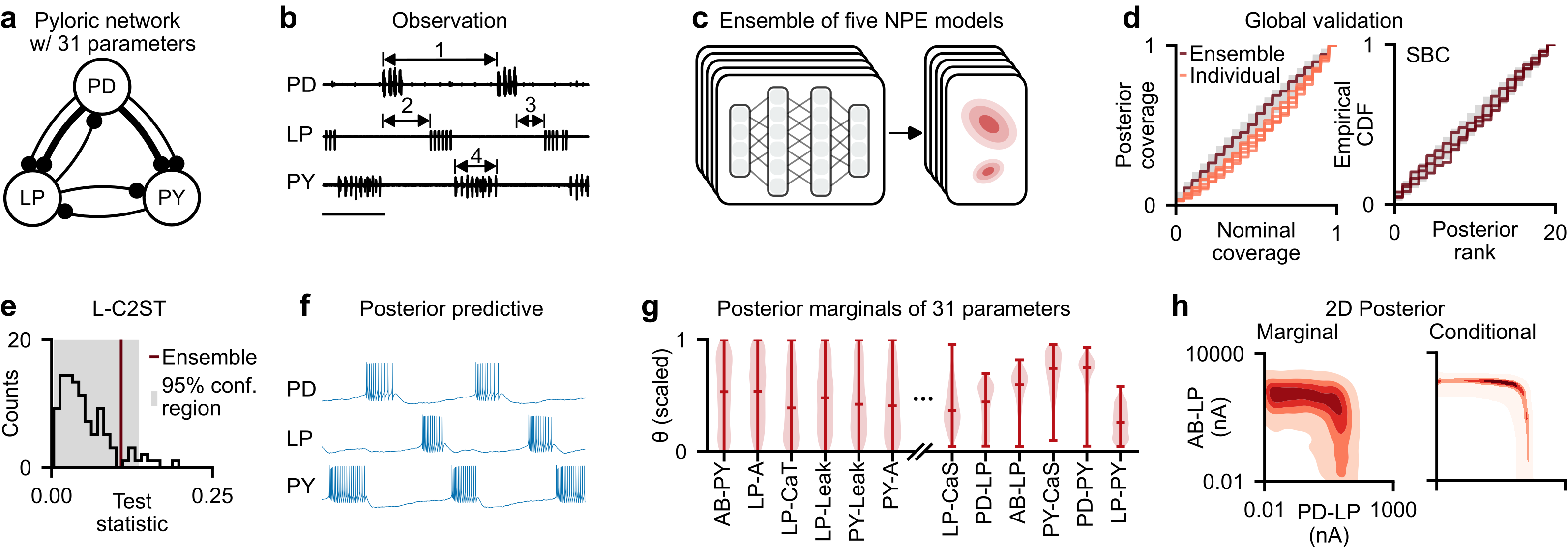}
    \caption{\textbf{Inference on neuroscience simulator with 31 parameters.
    (a)} Illustration of the pyloric network simulator. The simulator models voltage activity in three model neurons and has a total of 31 parameters.
    \textbf{(b)} Experimental voltage recording from this circuit. Arrows indicate expert defined summary statistics. Scale bar indicates 500\,ms.
    \textbf{(c)} We train an ensemble of five NPE models.
    \textbf{(d)} Global validation checks. Left: The ensemble has good expected coverage, whereas individual NPE models are overconfident. Right: The ensemble also has good marginal calibration.
    \textbf{(e)} Local C2ST. NPE ensemble (red) is within confidence region of test statistics (gray area).
    \textbf{(f)} Posterior predictive simulation matches summary statistics of experimental observation. Time scale as in panel (b).
    \textbf{(g)} 1D posterior marginal distributions, ordered by standard deviation (relative to prior).
    \textbf{(h)} Left: 2D posterior marginal for two of the parameters. Right: 2D posterior conditional for the same pair of parameters.
    }
    \label{fig:10_pyloric}
\end{figure}

Our third example is a biophysics simulator from neuroscience. The simulator models the neural activity in a small circuit from the crab \textit{Cancer borealis}. We will use this example to illustrate that SBI can scale to models with several dozens of parameters.

The simulator of the pyloric network consists of three neurons, each with eight parameters, and seven synapses, each with one parameter (Fig.~\ref{fig:10_pyloric}a). This leads to a total of 31 parameters. We aimed to infer these parameters based on 18 summary statistics of experimentally observed network activity (Fig.~\ref{fig:10_pyloric}b) \citep{prinz2004similar, haddad2018circuit, gonccalves2020training}. Following previous work, we used a uniform prior over parameters with bounds informed by previous studies \citep{prinz2004similar, gonccalves2020training}. We generated three million simulations by sampling from the prior and running the simulator. A challenge in this simulated dataset was that it contained many summary statistics that were far outside of the ranges of experimental recordings, and that many summary statistics were ill-defined. To avoid outliers, which could negatively impact neural network training, we capped all summary statistics to be within ranges that could be expected from experimental recordings. In addition, we replaced all undefined summary statistics with unreasonably large values (details in Appendix~Sec.~\ref{sec:A4_3_pyloric_details}), and used all three million simulations for training the inference network.

Due to the large dimensionality of the parameter space (and the associated larger cost of sampling from this space with MCMC or VI), we chose NPE as inference method. We used a neural spline normalizing flow as inference network. We trained an ensemble of five NPE models (each with a different random seed for neural network initialization and data loading), anticipating that individual networks might produce poorly calibrated posteriors for a simulator with many parameters (Fig.~\ref{fig:10_pyloric}c) \citep{hermans2022crisis}. Due to the large size of the training dataset (three million simulations), we used a large batchsize of 4096 and accelerated training on GPUs.

After training, we validated the trained NPE model (Fig.~\ref{fig:10_pyloric}d). First, we performed global validation with expected coverage. We found that each individual NPE model was slightly overconfident, but that the ensemble (i.e., the average posterior returned by the five models) showed good calibration. Similarly, the ensemble posterior also produced good marginal calibration, as evaluated with SBC. Next, we validated the posterior distribution for a specific experimental observation \citep{haddad2018circuit}. We used the local classifier-two-sample testing (L-C2ST) diagnostic (trained with 20k additional simulations), and found that the ensemble posterior also fell within the confidence region of the test statistic (Fig.~\ref{fig:10_pyloric}e). Finally, summary statistics of posterior predictive samples given this observation closely matched the experimental observation (Fig.~\ref{fig:10_pyloric}f).

Since no diagnostic tool indicated issues in the ensemble posterior, we continued our analysis. We first visualized the marginal distribution of the 31D posterior (Fig.~\ref{fig:10_pyloric}g). Similar to previous work \citep{prinz2004similar, gonccalves2020training}, we found that the marginals of the posterior were almost as broad as the uniform prior. This does, however, \emph{not} mean that experimental recordings do not inform the parameters. When inspecting the 2D posterior marginal, we found parameters that are tuned with respect to each other (Fig.~\ref{fig:10_pyloric}h, left). However, a 2D marginal distribution still integrates across parameters of all other 29 parameter dimensions \citep{gonccalves2020training, deistler2022energy, gao2024deep}. To gain further insight into the relative tuning of parameters, we inspected the conditional posterior distribution of two parameters, \emph{given a set of values for the other 29 parameters} (sampled from the posterior; Fig.~\ref{fig:10_pyloric}h, right). We found that this conditional 2D posterior distribution was very narrow and indicated strong co-tuning between these parameters. These results show that, even if the marginal distributions of the posterior may seem broad, individual parameters must still be carefully tuned to match experimental recordings.

Overall, this example demonstrates that NPE can scale inference to models with dozens of parameters and that an ensemble of NPE models can improve coverage. The example also shows that the resulting posterior can be used to study parameter sensitivities and the tuning of parameters relative to each other.

\section{Discussion}
\label{sec:5_discussion}

Statistical inference is at the core of many quantitative disciplines. Given observed data and candidates of models, it allows scientists and engineers to draw conclusions about the properties and mechanisms of the underlying system. 
However, many scientific models are implemented as numerical simulators, for which gradient-evaluations or likelihood evaluations are often intractable. 
This intractability makes statistical inference for simulators a significant challenge and has spurred the development of simulation-based inference (SBI).

For many decades, SBI methods were limited to Approximate Bayesian Computation approaches. However, these methods typically struggle to scale to high-dimensional data.
In recent years, driven by advances in probabilistic machine learning, neural-network-based methods for SBI have made it possible to overcome these limitations. These methods generate a dataset of parameters and simulated data by sampling from the prior and running the simulator, and then train a neural network on this dataset. After training, the neural network is used to perform inference for experimentally observed data. In the limit of infinite training data and a perfectly trained neural network, the resulting posterior estimate converges to the true posterior distribution. This approach is universally applicable because this workflow only requires simulations, with no need for likelihood or gradient evaluations.

Beyond enabling inference for any kind of simulator, SBI also has some advantages over traditional inference methods such as Markov-chain Monte-Carlo (MCMC): It can fully parallelize simulations and does not require \emph{any} sequential simulation steps, and it can amortize inference: After a costly initial phase of training and simulation, the posterior for new observations can be obtained almost instantly, enabling real-time and high-throughput applications.

Despite rapid progress over the past years, SBI methods still face several limitations: First, current SBI methods can still require many simulations. Neural networks can overfit or generalize poorly, especially in cases where too few simulations are used. Second, it can be difficult to obtain a posterior distribution that is \emph{exact}, even with many simulations \citep{lueckmann2021benchmarking}. The difficulty, however, depends on many factors such as data and parameter dimensions as well as model complexity.
Third, it has been demonstrated that neural-network-based SBI methods can be sensitive to misspecification \cite{cannon2022investigating, schmitt2023detecting, schmitt2024detecting, ward2022robust, kelly2024misspecification, huang2023learning, wehenkel2024addressing, gao2024generalized}. Therefore, to robustly perform inference on real-world observations, we will require methods that make the inference procedure more robust. Improving the simulation-efficiency, accuracy, and robustness of SBI methods are thus important directions for future research.

Beyond technical limitations, a significant practical hurdle for scientists is navigating the evolving SBI landscape. Practitioners are required to make critical choices regarding the specific algorithm, neural network architecture, training process, or sampling method, and appropriate diagnostic approaches to analyze the quality of the inference process and to identify potential problems. This paper aims to serve as a practical guide to bridge this gap. By focusing on a core set of commonly used methods, we hope to provide domain scientists with a deeper understanding of their internal workings and the associated trade-offs, making these tools more accessible. While this guide concentrates on this core set, we review several recent and advanced variants in Appendix~\ref{sec:A3_recent_developments} \citep{sharrock2022sequential, geffner2023compositional, dax2023flow, gloeckler2024allinone}.

In the past years, neural-network-based SBI methods have already been applied in a wide range of domains (Fig.~\ref{fig:A2_sbi_app_explorer}): 
For example, SBI has been used in neuroscience to infer conductances \citep{groschner2022biophysical, deistler2022energy}, plasticity rules \citep{confavreux2023meta,rossler2023skewed}, connectivity values \citep{myers2024disinhibition, boelts2023simulation}, and whole-brain dynamics \cite{jin2023bayesian,hashemi2023amortized}. In cognitive science, it has been used to fit models of perceptual decision making \citep{boelts2022flexible, von2022mental}; in biology to identify drivers of embryo-uterine interactions \citep{bondarenko2023embryo}, to infer models of population genetics \citep{korfmann2023deep}, and evolutionary parameters from adaptation dynamics \citep{avecilla2022neural}; and in the social sciences  to relate models of individual behavior to aggregate data \citep{dyer2022black, ciganda2025}. 
In physics, it has been used to characterize Galactic Center $\gamma$-ray excess \citep{mishra2022neural}, to infer parameters of black hole and binary neutron star mergers given gravitational wave recordings \citep{dax2021real, dax2025real},  to amortize inference in spectral energy distribution models \citep{hahn2022accelerated}, to infer cosmological parameters from field-level analysis of galaxy clustering \citep{lemos2024field}, and to characterize the properties of exoplanets \cite{Barrado2023, Gebhard_2025_Flow}.

Machine learning has revolutionized many parts of science. Recent simulation-based inference methods powered by neural networks enable inference on problems that were previously inaccessible. Because of the broad applicability of SBI, we expect that it will become a fundamental tool for domain scientists across many fields, and that it will enable new scientific discoveries. 

\section*{Code availability}
We provide code to reproduce all results and examples at \url{https://github.com/sbi-dev/sbi-practical-guide}. We used the \texttt{sbi} toolbox to implement the examples \citep{boelts2024sbi}. The \texttt{sbi} toolbox implements many popular algorithms for simulation-based inference, as well as many diagnostics (e.g., SBC, expected coverage, TARP, L-C2ST), analysis tools (e.g., visualization, marginal and conditional moments, or MAP-estimation), and utilities (e.g., building an ensemble of posteriors). The \texttt{sbi} toolbox is available at \url{https://github.com/sbi-dev/sbi}, documentation is available at \url{https://sbi.readthedocs.io/en/latest/}.

\section*{Acknowledgements}
We thank all members of the involved research groups as well as the entire SBI community for valuable discussions and input.
This work has been supported by
the German Federal Ministry of Education and Research  (Tübingen AI Center FKZ: 01IS18039A),
the German Research Foundation (DFG) through Germany’s Excellence Strategy (EXC-Number 2064/1, PN 390727645) and SFB1233 (PN 276693517), SPP 2041 (PN 34721065),  SPP 2298-2 (PN 543917411), SFB 1233 `Robust Vision', the  Carl Zeiss Foundation  (Certification and Foundations of Safe Machine Learning Systems in Healthcare)
and the European Union (ERC, ``DeepCoMechTome'', ref. 101089288).
BKM was part of the ELLIS PhD program during this work, receiving travel support from the ELISE mobility program which has received funding from the European Union’s Horizon 2020 research and innovation programme under ELISE grant agreement No 951847.
TM and PLCR were supported from a national grant managed by the French National Research Agency (Agence Nationale de la Recherche) attributed to the ExaDoST project of the NumPEx PEPR program, under the reference ANR-22-EXNU-0004.
PS is supported by the Helmholtz Association Initiative and Networking Fund through the Helmholtz AI platform grant.
JL is recipient of the Pierre-Aguilar Scholarship and thankful for the funding of the Capital Fund Management (CFM).
MD, MG, GM and JKL are members of the International Max Planck Research School for Intelligent Systems (IMPRS-IS).


\bibliography{refs.bib}

\clearpage
\appendix
\setcounter{figure}{0}
\renewcommand{\thefigure}{A\arabic{figure}}
\setcounter{section}{0}
\renewcommand{\thesection}{A\arabic{section}}

\section{Choosing the inference method: NPE, NLE, NRE, and beyond}
\label{sec:A1_choosing_a_method}

\begin{figure}[h!]
    \centering
    \includegraphics[]{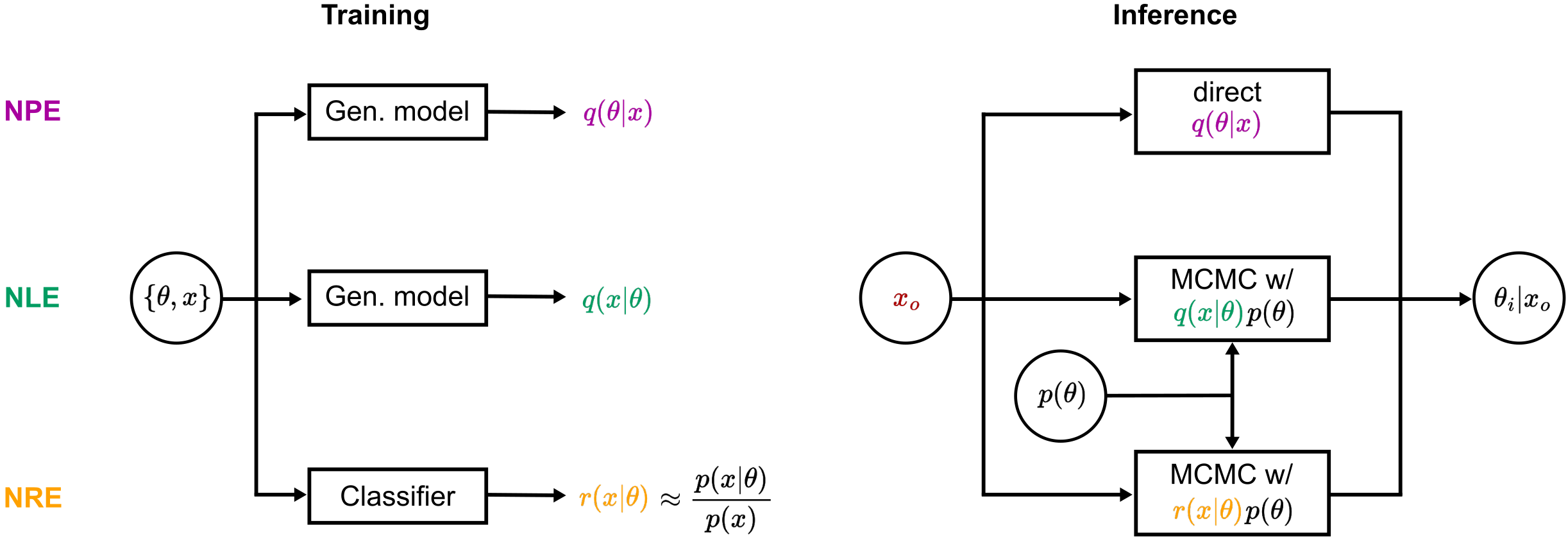}
    \caption{\textbf{Overview of SBI methods.} For NPE and Neural Likelihood Estimation (NLE) a conditional generative model (Gen.~model) is trained, whereas for Neural Ratio Estimation (NRE) a classifier is trained (left).
    During inference time the trained network in NPE can be evaluated directly, while for NLE and NRE an additional inference step via MCMC is necessary (right).}
    \label{fig:sbi_methods}
\end{figure}

Over the past years, a broad range of new SBI methods have been developed. All of these methods share the commonality of training on simulated data and then evaluating the trained network on the observation to obtain the posterior. Still, the methods differ in many important aspects, such as their ability to use embedding networks, to deal with high-dimensional observations or i.i.d.~observations, their training cost, and their inference speed and accuracy. Figure~\ref{fig:sbi_methods} provides a conceptual overview of the three core methods---NPE, NLE, and NRE---illustrating their distinct training targets (conditional generative models for NPE/NLE, classifier for NRE) and the mechanisms by which they perform inference (direct evaluation for NPE vs. additional MCMC/VI steps for NLE/NRE). This section provides detailed guidance on the differences, serving as a recipe for choosing the most suitable SBI algorithm for the research problem at hand. We expand on the brief overview provided in Section~\ref{sec:3_2_choosing_components} with practical considerations, detailed comparisons, and recent developments. As new SBI methods are developed regularly, this section might not reflect the future state of SBI methods, however underlying principles will persist.

\subsection{Considerations for choosing an SBI method}
\label{sec:A1_1_considerations_for_choosing}

Choosing the most appropriate SBI method involves weighing several factors related to the problem characteristics and desired inference properties. We provide a short summary of these factors for each of the SBI core methods in Table \ref{tab:method_comparison}. In the following, we give detailed guidance on each factor to help make informed decisions. Section \ref{sec:A1_2_summary_and_recommendations} synthesizes these considerations into concrete recommendations for selecting a method.

\paragraph{Inference speed and sampling cost}\label{par:inference_speed}
This is often an important distinction between methods. NPE directly predicts the posterior distribution which, after training, allows for very  fast inference: Obtaining posterior samples typically takes milliseconds through a single forward pass of the network. This speed is crucial for applications requiring real-time analysis or when performing inference on thousands of observations (e.g., analyzing data from many experimental subjects).

In contrast, NLE and NRE require subsequent MCMC or VI steps to sample from the posterior, which are computationally more costly. While these sampling steps use the trained neural network for efficient evaluations, they can still take several seconds or minutes, especially for models with many parameters (Fig.~\ref{fig:5_mcmc_or_vi_failures}a). Computational cost scales roughly as $\mathcal{O}(d \cdot n_{\text{samples}} \cdot n_{\text{chains}})$ for MCMC, where $d$ is the parameter dimension. Moreover, the reliance on an external sampling algorithm introduces an additional source of potential error (e.g., poor mixing of MCMC chains, mode collapse in VI, Fig.~\ref{fig:5_mcmc_or_vi_failures}b). 

\textbf{Practical guidance:} If you need to perform inference multiple times (e.g., more than 100 times) or require sub-second inference, strongly consider NPE. For one-off analyses where inference time is not critical, NLE/NRE remain viable options.

\begin{table}[t]
\centering
\begin{tabular}{l|c|c|c}
\hline
\textbf{Criterion} & \textbf{NPE} & \textbf{NLE} & \textbf{NRE} \\
\hline
Inference speed & ++ (ms) & $-$ (sec-min) & $-$ (sec-min) \\
Simulation efficiency for i.i.d. data & $--$ & ++ & ++ \\
Embedding network compatibility & ++ & $--$ & ++ \\
Training computational cost & Medium & Medium & Low \\
Invalid simulation handling & ++ & $-$ & $-$ \\
Hyperparameter tuning complexity & Low & High & High \\
\hline
\end{tabular}
\caption{Comparison of SBI methods. ++ indicates strong advantage, + moderate advantage, $-$ moderate disadvantage, $--$ strong disadvantage. Hyperparameter tuning is more complex for NLE/NRE due to additional MCMC/VI parameters.}
\label{tab:method_comparison}
\end{table}

\paragraph{Data dimensionality and embedding networks}\label{par:embeddings}
The architectural differences between methods have implications for handling high-dimensional data. NPE and NRE both take observed data $\bx$ as \emph{input} to their neural networks, allowing them to work with high-dimensional simulations, e.g. by using embedding networks—specialized architectures designed to extract relevant features from complex data types. Common embedding networks include:
\begin{itemize}
    \item Convolutional Neural Networks (CNNs) for image data,
    \item Recurrent Neural Networks (RNNs) or Transformers for time series,
    \item Graph Neural Networks (GNNs) for relational data.
\end{itemize}
These embedding networks can be trained end-to-end with the generative model or classifier, allowing gradients to flow through the entire architecture and automatically learning most informative features for inference. NLE, however, takes parameters $\btheta$ as input to predict data $\bx$, fundamentally limiting its ability to leverage such embedding networks. While workarounds exist (e.g., pre-computing fixed embeddings/summary statistics), they typically underperform end-to-end learning.

\textbf{Practical guidance:} For high-dimensional raw data (images, time series, graphs), use NPE or NRE. Only consider NLE if your data is low-dimensional, or you  have  validated summary statistics.

\paragraph{Amortized or sequential inference}
Inference methods such as NPE, NLE, or NRE are amortized: After an initial phase of simulation and training, they can perform inference for any observation. This can be beneficial when inference is performed for many observations.

When inference is only performed for one, or few, observations, then \emph{sequential} methods can largely improve simulation efficiency of inference methods \citep{lueckmann2021benchmarking}. These methods perform inference across several rounds and draw parameter sets from a distribution that is different from the prior. We provide additional background and guidance on these methods in Appendix~Sec.~\ref{sec:A2_sequential_methods}.

\textbf{Practical guidance:} Use amortized methods when inference is to be performed for many observations. For few observations, consider sequential methods if simulation efficiency is a bottleneck.

\paragraph{Datatype: i.i.d. observations}\label{par:iid_observations}
The structure of observed data fundamentally affects method efficiency. Many experimental settings produce multiple independent and identically distributed (i.i.d.) observations—for example, multiple trials from a single participant in a psychology experiment, or multiple measurements from the same experimental condition.

NLE and NRE naturally excel in these settings because they can be trained on single observations, then efficiently combine evidence during inference based on $N$ observations:
\begin{equation}
p(\btheta | \bx_1, \ldots, \bx_N) \propto \prod_{i=1}^N p(\bx_i | \btheta) p(\btheta) \quad \text{(for NLE)}.
\end{equation}

NPE, conversely, faces significant challenges with i.i.d. data. The network must be trained on datasets where each parameter set is paired with multiple simulations (e.g., variations of one up to the maximum number expected during inference) and include permutation invariant components \cite{zaheerDeepSets2018,radev2020bayesflow} which act on a set of observed trials. These requirements increase simulation cost by a factor of $N_{\max}$, require specialized permutation-invariant architectures and limit flexibility if experimental designs change.

\textbf{Practical guidance:} For i.i.d. observations, prefer NLE or NRE unless you need amortized inference across many different datasets. Works on exchangeable neural networks \citep{chan2018exchangeable,radev2020bayesflow} and score-based methods (see below, \cite{linhart_diffusion_2024c}) address NPE's limitations but add implementation complexity.

\paragraph{Training cost}\label{par:training_cost}
The computational cost of training varies between methods due to their different architectures and objectives. NPE and NLE typically train conditional generative models (normalizing flows, diffusion models), which require computing and inverting Jacobians (for flows), multiple forward/reverse passes (for diffusion models) and overall careful numerical stability considerations.

In contrast, NRE trains a simpler binary classifier, typically requiring 2-3 times fewer FLOPs per training step.

\textbf{Practical guidance:} Training cost rarely dominates the computational budget and is therefore typically not the main criterion for the method choice, unless one is extremely compute-constrained.

\paragraph{Handling invalid simulations}\label{par:invalid_sims}
Real-world simulators often fail for certain parameter combinations---numerical instabilities in neuroscience or climate simulators, non-physical configurations in molecular simulations, or convergence failures in optimization-based simulators. SBI methods differ in their ability to handle such failures.

NPE can simply discard invalid simulations during data generation, as it learns $p(\btheta|\bx)$ only from valid parameter-data pairs. For extreme cases with high proportions of invalid simulations, there have been proposed methods to train an additional classifier that predicts the success of a simulation beforehand in order to save the simulation budget \cite{lueckmann2017flexible, deistler2022energy}. 
NLE and NRE, however, need to explicitly model invalid simulations as well, e.g., by assigning special values (e.g., extreme likelihoods) to failed simulations or training separate models for the failure probability, which complicates implementation and potentially reduces performance. If invalid simulations are simply discarded for NLE or NRE, then the resulting posterior estimate can be systematically biased towards parameter regions that often generate failed simulations \citep{gloeckler2022variational}.

\textbf{Practical guidance:} If simulator failures are more common (for > 1\% of parameters) and clustered in parameter space, we recommend NPE and its variants with additional classifiers\footnote{see e.g.\href{https://sbi.readthedocs.io/en/latest/advanced_tutorials/06_restriction_estimator.html}{this user guide} in the \texttt{sbi} package for a user-friendly implementation}. 

\paragraph{Hyperparameter tuning requirements}\label{par:hyperparameters}
For the neural network architecture, systematic hyperparameter optimization can be performed for all methods, although software packages like the \texttt{sbi} toolbox offer good defaults. Common approaches include random search or Bayesian optimization over architecture choices (number of layers, hidden units, flow transformations) and training hyperparameters (learning rate, batch size, weight decay). Tools like MLFlow \cite{zaharia2018accelerating}, Weights \& Biases \cite{wandb}, or Hydra \cite{yadan2019hydra} can automate this process. The key is selecting an appropriate validation metric: For NPE, minimizing validation loss (negative log-probability under the estimated posterior) provides a principled objective, as this loss upper bounds the Kullback-Leibler divergence between the true and approximate posterior \citep{lueckmann2021benchmarking}. Similarly, for NLE, the validation loss bounds the KL divergence between true and approximate likelihoods. For NRE, the binary cross-entropy loss serves as the validation metric, though its relationship to posterior quality is less direct.

Beyond the neural network hyperparameters, NLE and NRE introduce additional complexity through their MCMC or VI components. Important choices include:
\begin{itemize}
    \item \textbf{MCMC}: Sampler type (HMC, NUTS, Random Walk), step sizes, number of chains, warmup steps, thinning,
    \item \textbf{VI}: Variational family, optimization algorithm, convergence criteria, initialization strategy.
\end{itemize}

Poor choices can lead to biased posteriors even with perfectly trained neural networks (Fig.~\ref{fig:5_mcmc_or_vi_failures}). NPE avoids this entirely---once trained, posterior sampling requires no additional hyperparameters.

\textbf{Practical guidance:} Budget significant time for MCMC/VI tuning with NLE/NRE. Consider using automated tools like ArviZ \cite{arviz_2019} to diagnose potential MCMC sampling failures. With NPE, focus tuning efforts solely on the neural network. For all methods, use held-out validation data to guide architecture choices, as overfitting to training data can produce overconfident posteriors.

\subsection{Explicit recommendations}
\label{sec:A1_2_summary_and_recommendations}

Based on the considerations above, we provide the following practical recommendations. These recommendations should be interpreted as simple heuristics, and might, of course, not be adequate for specific problems. 

\textbf{NPE as default starting point: } Begin with NPE using normalizing flows unless your specific use case clearly favors another method. NPE's combination of fast inference (\S\ref{par:inference_speed}), implementation simplicity, and robust handling of simulator failures (\S\ref{par:invalid_sims}) makes it suitable for most applications.

\textbf{When to use NLE:} Choose NLE primarily for i.i.d. observation settings (\S\ref{par:iid_observations}) where (1) simulation is expensive, (2) the number of observations varies significantly, and (3) you can afford the computational cost of MCMC (\S\ref{par:inference_speed}). Common applications include cognitive modeling, population genetics with multiple individuals, or repeated measurements in physics experiments.

\textbf{When to use NRE:} Consider NRE for similar settings as NLE but when training computational budget is highly constrained (\S\ref{par:training_cost}). The simpler classifier architecture can be advantageous for very high-dimensional parameter spaces where generative modeling becomes challenging.

\textbf{When to consider sequential methods:} If you have expensive simulations and need inference for a single observation, sequential variants (Appendix~Sec.~\ref{sec:A2_sequential_methods}) (SNPE, SNLE, SNRE) can reduce simulation costs substantially. However, these methods require careful monitoring for convergence issues and introduce additional hyperparameters that need tuning for optimal performance.

\textbf{When to consider emerging methods:} Consider score-based methods or flow matching (Appendix Sec.~\ref{sec:A3_recent_developments}) when you need the flexibility of NPE with better i.i.d.~handling, and can accept slower inference. Tabular foundation models can improve inference for low simulation budgets. The Simformer improves the flexibility of inference by estimating the likelihood and posterior, and by handling missing or unstructured data.

\textbf{Validation through multiple methods:} When possible, implement multiple methods for important applications. Agreement between NPE and NLE/NRE provides evidence for reliable inference, while disagreement can reveal issues with simulator design, prior specification, or implementation errors.

We stress that implementation quality often dominates method choice---a well-tuned NPE will outperform a poorly configured NLE, even in cases where NLE has potential theoretical advantages. Invest time in hyperparameter optimization, architecture design, and diagnostic checking regardless of method choice.

\section{Sequential methods for targeted inference}
\label{sec:A2_sequential_methods}

When inference is needed for a single specific observation rather than amortized inference across many observations, sequential methods can dramatically improve efficiency \cite{papamakarios2016fast,lueckmann2017flexible,greenberg2019automatic,papamakarios2019sequential,hermans2020likelihood}. These methods, also called active learning approaches \cite{thomas2022likelihood,lueckmann2019likelihood}, iteratively refine the inference by focusing simulations on parameter regions most relevant to the observed data.

Sequential methods operate in multiple rounds:
\begin{enumerate}
    \item In the initial round: Sample parameters from the prior, run simulations, train the neural network.
    \item In subsequent rounds: Use the current posterior estimate or other proposal distribution to propose new parameters, simulate these targeted parameters, retrain the inference network on the combined dataset.
    \item Repeat until convergence or computational budget exhausted.
\end{enumerate}
This focused sampling can reduce simulation requirements by an order of magnitude or more \cite{lueckmann2021benchmarking,gloeckler2022variational}, particularly when the prior is broad relative to the posterior. In extreme cases with uninformative priors and precise data, sequential methods may be necessary to achieve any meaningful inference \citep{deistler2022truncated}.

\paragraph{Key advantages}
Sequential methods can be particularly beneficial when dealing with expensive simulations that take minutes to hours per run, making it crucial to focus computational resources on the most relevant parameter regions. They also excel in scenarios with broad priors—where prior uncertainty spans many orders of magnitude but the posterior is expected to be concentrated in a small region. The approach is especially valuable when highly accurate inference is needed for a single specific observation, justifying the computational investment in targeted sampling. 

\paragraph{Implementation challenges and solutions}
While powerful, sequential methods introduce additional challenges:

\textbf{Training instabilities:} Early sequential methods (particularly sequential NPE variants) required modifying the loss function to account for the non-prior proposal distribution, often leading to numerical instabilities. More recent approaches like truncated SNPE \cite{deistler2022truncated} avoid this by sampling from a truncated prior, maintaining the standard loss function and training stability.

\textbf{Loss of amortization:} The trained network becomes specialized for one specific observation, losing the ability to perform inference on new data without retraining. This trade-off is acceptable when inference is needed for a single critical observation but problematic for applications requiring repeated inference.

\textbf{Diagnostic limitations:} Diagnostics (e.g., global simulation-based calibration as well as local diagnostic tools) become computationally prohibitive as they would require running the sequential procedure for many test observations. Alternative validation strategies are typically necessary \citep{deistler2022truncated}.

\textbf{Hyperparameter choices:} All sequential methods come with additional hyperparameter choices like the number of inference rounds, number of simulations per round, and proposal strategies.

\paragraph{Practical recommendations for sequential methods}

When implementing sequential methods, begin with non-sequential approaches to establish a baseline and verify your simulator and prior specification work correctly. Throughout the sequential procedure, carefully monitor convergence by tracking how the posterior evolves across rounds---stability in the posterior shape indicates convergence. Crucially, ensure sufficient simulations in the initial round to obtain a reasonable first posterior approximation; too few initial simulations can misguide the focusing process, causing the method to concentrate on incorrect parameter regions from which it may never recover. While it can be tempting to distribute simulations evenly across rounds, investing more heavily in early rounds often yields better results than spreading simulations too thinly.

Sequential variants exist for all major methods (SNPE, SNLE, SNRE), with similar performance characteristics. The choice between them typically follows the same considerations as their non-sequential counterparts.

\section{Recent developments and emerging methods}
\label{sec:A3_recent_developments}

The field of SBI is rapidly evolving, with new methods emerging regularly to address limitations of the core NPE, NLE, and NRE approaches. Here we highlight several promising developments that offer novel solutions to common challenges such as handling i.i.d.~data, improving sample quality, or eliminating training altogether. While this is not an exhaustive list, these methods represent recent advances that can be useful when standard approaches fall short.

\paragraph{Improved generative models}

A central line of improvement for SBI methods over the past years has been to improve the generative models used as inference networks in NPE.

Firstly, recent work has introduced score-based methods that learn the score function $\nabla_{\btheta} \log p(\btheta|\bx)$ rather than the posterior density itself \citep{geffner2023compositional, sharrock2022sequential}. These methods work by training a neural network to predict the score at different noise levels, then using this for sampling via Langevin dynamics or more sophisticated SDE solvers.
These methods enable the usage of the expressive powers of diffusion models for NPE. Additionally, they have the ability to handle i.i.d. observations naturally \citep{geffner2023compositional,linhart_diffusion_2024c}, similar to NLE/NRE, while maintaining NPE's architectural flexibility. The score factorizes for independent observations:
\begin{equation}
\nabla_{\btheta} \log p(\btheta | \bx_1, \ldots, \bx_N) = \nabla_{\btheta} \log p(\btheta) + \sum_{i=1}^N \nabla_{\btheta} \log p(\bx_i | \btheta)
\end{equation}
However, these methods require solving SDEs or ODEs during sampling, making them slower than standard NPE (typically seconds rather than milliseconds). They represent a middle ground: More flexible than NLE/NRE but more capable with i.i.d. data than standard NPE.

Secondly, flow matching \citep{lipman2023flow, dax2023flow,moss_fnope_2025} has emerged as a powerful alternative to traditional normalizing flows for NPE. Instead of constructing flows through composed bijections with tractable Jacobians, flow matching learns a continuous-time normalizing flow by regressing onto a pre-specified conditional vector field. Key advantages include: No architectural constraints---any neural network can parameterize the vector field, which can lead to faster and more stable training; more expressive power leading to often superior sample quality compared to discrete time normalizing flows.
The main drawback is sampling speed: Generating samples requires solving an ODE (typically 50-100 neural network evaluations), compared to a single pass for discrete flows. This makes flow matching ideal when sample quality is paramount but inference speed is less critical.

To improve sampling speed, \citet{schmitt2024consistency} suggested to use consistency models for NPE. These models can improve sampling speed, but they do not easily provide density evaluations, and they might require tuning of an additional hyperparameter for the number of sampling steps.

Thirdly, tabular foundation models have been reported to perform well for NPE \citep{vetter2025effortless}. These methods use general-purpose probabilistic foundation models, such as TabPFNv2~\citep{hollmann2025accurate}, to estimate the posterior. This approach does not require any training phase and these models can perform particularly well for small training datasets (i.e., low simulation budgets). However, this convenience comes at the cost of slower inference speed and potentially lower accuracy when very large simulation budgets are available.

\paragraph{Improving robustness to model misspecification}

It has been demonstrated that SBI methods based on neural networks can show poor performance when applied to misspecified observations \citep{cannon2022investigating}. Over the past years, improving robustness to model misspecification has become a popular research topic. These methods either aim at detecting misspecification \citep{schmitt2023detecting, schmitt2024detecting}, or at performing inference despite such misspecification. In the latter case, it has been proposed to add noise to simulations \citep{ward_robust_2022}, to learn robust statistics \citep{huang2023learning}, to modify the inference task \citep{gao2024generalized}, or to use unlabeled \citep{mishra2025robust} or labeled data \citep{wehenkel2024addressing} during training.

\paragraph{Improving the flexibility of inference}

A limitation of standard SBI methods such as NPE is that they have limited flexibility in handling missing, unstructured, or function-valued data. To improve the flexibility of SBI, the Simformer \citep{gloeckler2024allinone} learns the full joint distribution $p(\btheta, \bx)$ with a single neural network model. Built on a transformer architecture with diffusion-based generative modeling, it trains on various combinations of observed and unobserved parameters and data. At inference time, this unified model can perform multiple tasks by masking different variables. For posterior inference $p(\btheta | \bx)$, the data $\bx$ is marked as observed while parameters $\btheta$ are estimated with the generative model. For synthetic data generation or likelihood evaluation via $p(\bx | \btheta)$, the parameters are observed while data is estimated with the generative model.

The architecture handles missing data and varying numbers of observations naturally through attention masking. For i.i.d.~data, it can process arbitrary numbers of observations without retraining, unlike standard NPE. The transformer's self-attention mechanism learns relationships between parameters and data flexibly, potentially capturing complex dependencies that task-specific architectures might miss.

Another powerful capability enabled by this joint modeling is generating posterior predictive distributions without running the simulator. Given observed data $\bx_o$, one can sample parameters $\btheta \sim p(\btheta | \bx_o)$ from the posterior, then directly sample new observations $\tilde{\bx} \sim p(\bx | \btheta)$ from the learned model. This provides full uncertainty quantification for predictions, incorporating both posterior uncertainty in parameters and the inherent stochasticity of the data-generating process---all without additional simulator calls.

Current limitations include higher computational requirements (both training and inference) and the need for careful positional encoding design. However, for applications requiring multiple types of inference or flexible handling of data structures, the Simformer's versatility may outweigh these costs.

\section{SBI applications database: Data collection and interactive explorer}
\label{sec:A5_sbi_database}

\begin{figure}[t]
\centering
\includegraphics[width=\textwidth]{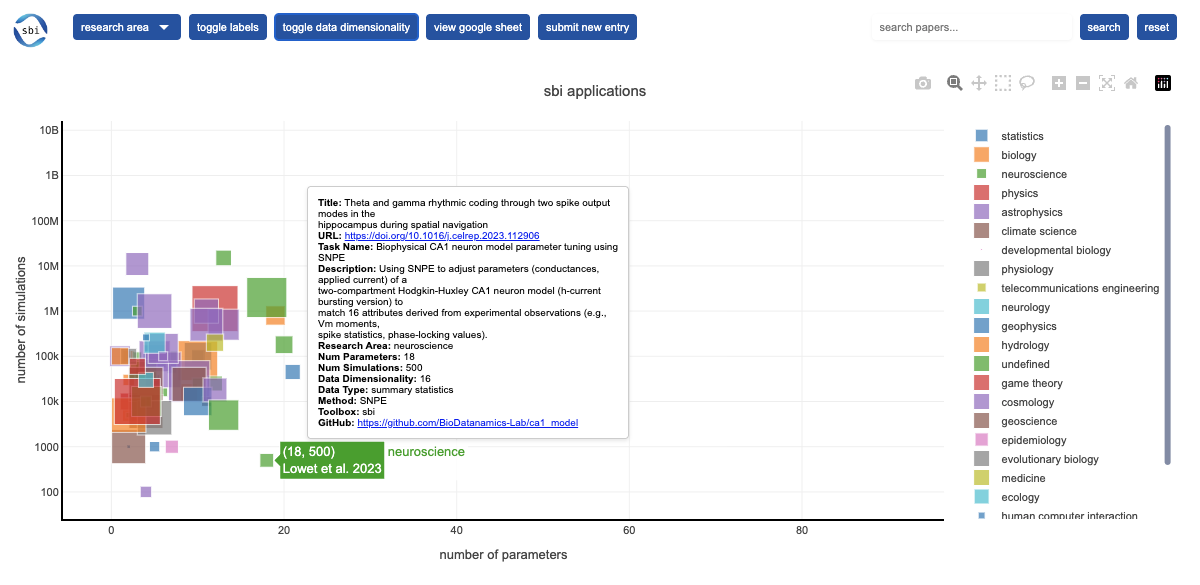}
\caption{\textbf{Interactive SBI Applications Explorer interface.} The web application displays published SBI applications as an interactive scatter plot showing the relationship between number of simulator parameters (x-axis) and number of training simulations used (y-axis, log scale). Points are colored by research domain (legend on right), with marker size proportional to data dimensionality. Hovering over individual points reveals detailed metadata including paper title, implementation details (SBI method, software toolbox), and links to publications and code repositories. The interface provides multiple filtering options (top navigation bar) allowing users to explore applications by research area, toggle visualization features, access the underlying database via Google Sheets, and submit new entries. In this example, hovering over a neuroscience application (Lowet et al., 2023) shows it used SNPE with 500 simulations for an 18-parameter hippocampal simulator given 16-dimensional summary statistics.}
\label{fig:A2_sbi_app_explorer}
\end{figure}

To provide practitioners with a comprehensive resource for exploring real-world SBI applications, we developed a curated database and interactive web application (see Fig.~\ref{fig:A2_sbi_app_explorer}). This effort complements existing resources in the SBI community, most notably the \url{simulation-based-inference.org} website, which provides an automated, continuously updated collection of SBI papers across all methodological and application domains \cite{sbi_org_website}. While that platform provides comprehensive coverage through automated paper collection, our database focuses on manually extracting practical implementation details that might be important for practitioners planning their own SBI projects.

\paragraph{Data collection methodology}
Our data collection process employed several complementary approaches:
\begin{itemize}
\item \textbf{Literature tracking}: We maintained a dedicated Zotero collection, systematically adding papers that cited the \texttt{sbi} toolbox or described SBI applications using other implementations.
\item \textbf{Algorithmic discovery}: We leveraged machine learning-based recommendation systems from Google Scholar and Scholar Inbox to identify relevant papers beyond direct citations, capturing applications using custom implementations or alternative toolboxes.
\item \textbf{Feature extraction}: We carefully designed a taxonomy of SBI-relevant features through iterative refinement, establishing clear criteria for each field to ensure consistency across different scrapers. Key features include simulation budget, parameter dimensionality, data characteristics, implementation details, and domain-specific metadata.
\item \textbf{Quality control}: While we employed large language models to assist in initial information extraction, every entry underwent careful human verification to ensure accuracy. Multiple researchers participated in the scraping process, with standardized protocols for handling ambiguous cases.
\end{itemize}

We prioritized extracting the most practically relevant information---particularly the number of simulations used and parameter dimensionality---while acknowledging that some implementation details may be unavailable or ambiguously reported in publications. Our database tracks individual inference tasks rather than papers, recognizing that single publications often demonstrate multiple distinct applications.

\paragraph{Scope and limitations}
This database is explicitly \emph{not} a systematic review, but rather a community resource designed to grow and improve over time. We acknowledge several limitations:
\begin{itemize}
\item Coverage bias toward applications using the \texttt{sbi} toolbox, though we actively sought implementations using other packages
\item Potential gaps in rapidly evolving fields where preprints and conference papers may not yet be indexed
\item Variations in reporting standards across disciplines affecting data completeness
\end{itemize}

We view these limitations not as flaws but as opportunities for community contribution. The database serves as a starting point---a foundation that the SBI community can collectively build upon.

\paragraph{Interactive explorer and community contributions}
The web application (\url{https://sbi-applications-explorer.streamlit.app/}) transforms this static database into a dynamic exploration tool. Users can filter applications across multiple dimensions, discover relevant prior work, and identify common practices within their domain (Fig.~\ref{fig:A2_sbi_app_explorer}). Recognizing that the SBI landscape evolves rapidly, we implemented mechanisms for community participation:
\begin{itemize}
\item Integrated \href{https://docs.google.com/forms/d/e/1FAIpQLSeu7Er272IKAnTxBX6osqcbrdvG2ny-aIybv6FDIFLLe8SSoA/viewform}{submission form} for adding new applications or correcting existing entries
\item Transparent review process with submissions collected via Google Forms and reviewed before integration
\item Open-source codebase (\url{https://github.com/sbi-dev/sbi-app-explorer}) enabling community contributions to functionality
\item Commitment to ongoing maintenance and regular updates
\end{itemize}

We encourage all SBI practitioners to contribute their applications, ensuring this resource remains comprehensive and current. Whether adding a missing paper, correcting metadata, or suggesting new features, community engagement is essential for maximizing value to the field.
\section{Diagnostic tools}
\label{sec:A4_diagnostic_tools}

This section provides detailed descriptions of the coverage diagnostic methods introduced in the main text. We explain both global diagnostics that assess posterior calibration on average across observations, and local diagnostics that evaluate calibration for specific individual observations.

\paragraph{Global diagnostic tools}
SBC, expected coverage, and TARP compare the obtained posterior distributions to the \emph{ground truth} parameter sets $\btheta$ (that were used to generate the simulation outputs). They do so by ranking the \emph{ground truth} parameter set $\btheta$ among $M$ posterior samples $\theta^q_1, \dots, \theta^q_M$ (obtained via the considered SBI method for the corresponding simulation output):
\begin{equation}
    r(\{f(\theta^q_1), \dots, f(\theta^q_M)\}, f(\theta)) = \sum_{m=1}^M \mathbb{I}(f(\theta^q_m) < f(\theta)) \in [0, M]
\end{equation}
where the rank $r$ is defined for a given univariate projection
$f: \theta \mapsto f(\theta) \in \mathbb{R}$.
They perform this comparison in different ways: SBC computes the ranks for each marginal separately,
without taking into account the possible correlations between parameters (in this case $f$ outputs each univariate coordinate of the parameter set separately). Expected coverage on the other hand computes, for every posterior sample and for the ground truth parameters, the (unnormalized) log-probability under the SBI posterior: $f(\theta) = \log q(\theta \mid x)$. Intuitively, expected coverage evaluates whether the ground truth parameters are with $k$\,\% probability in the $k$\,\% confidence region of the posterior, also known as the (expected) coverage. This allows for a more straightforward and efficient diagnostic procedure, especially in high dimensional parameter spaces. 

Tests of Accuracy with Random Points (TARP) \cite{lemos2023sampling} provides an alternative approach that does not require posterior density evaluations. Instead of using highest posterior density regions, TARP constructs credible regions as balls centered at randomly chosen reference points throughout the parameter space, with radii extending to include the ground truth parameters. By varying the reference point distribution and checking coverage across different credibility levels, TARP can detect posterior inaccuracies that other methods might miss, making it particularly useful for SBI approaches based on  generative models where explicit density evaluations are infeasible.

Finally, to diagnose the calibration of the SBI posterior, these methods generate a histogram of the computed ranks across all $N$ posteriors. As shown by \citet{cook2006validation}, this histogram must be uniformly distributed between [0, M]. If this is not the case, then inference is, on average, flawed. This can be visualized on so-called P-P plots (Fig.~\ref{fig:7_diagnostics}b), where any deviance from the diagonal (black dashed lines) indicates a mismatch with the uniform rank distribution, hence a flawed SBI posterior. A significant advantage of these plots is that they are easy to interpret and can indicate the nature of the failure mode, e.g. whether the posterior is too narrow, too broad, or skewed towards too low or too high parameters.

\paragraph{Local diagnostic tools}
Local diagnostic tools can check calibration for individual observations. Below, we provide details on Local Coverage Test (LCT) \cite{zhao2021diagnostics} and Local Classifier-two-sample testing (L-C2ST) \cite{linhart2024c2st}. Both methods rely on an additional simulated calibration dataset, used to train a regression model that learns statistical quantities as a function of the observation space, that are then used to quickly (in an amortized way) diagnose the SBI posterior for any new observation. Typically, SBC or HPD require between 1000 and 10000 simulations to give reliable results.

LCTs are a direct extension of expected coverage tests, where the coverage is now computed \emph{conditionally} on the observation space, allowing diagnosis of the SBI posterior for a specific (measured) observation. The idea is to again use samples from the prior and corresponding simulated observations and regress the rank statistic on the observation space (in other words, the projection $f$ is now defined for every $\theta$ \emph{and} $x$). Any regression model (random forest, MLP, etc.) can be used. This allows direct learning of the coverage as a function of $x$, that can then be plotted to quickly diagnose the SBI posterior for every new observation $\bx_o$.

L-C2ST is a more recent local validation method proposed by \citet{linhart2024c2st} to address the limitations of LCTs. Based on the well-known classifier two-sample test (C2ST) \cite{Lopez2016}, the idea is to train a binary classifier to distinguish between the true and estimated posterior. As no samples from the true posterior are available, the idea is to train the classifier on samples from the \emph{joint distributions}:
\begin{equation}
     (\btheta_n, \bx_n)\mid (C=0) \sim p(\btheta, \bx) \quad \mathrm{vs.} \quad(\btheta^q_n, \bx_n) \mid (C=1) \sim q(\btheta \mid \bx)p(\bx)~.
\end{equation}
Similarly to coverage tests, they are obtained by first sampling parameters from the prior, generating corresponding observations via the simulator and then inferring the associated SBI posterior samples.
This way, L-C2ST implicitly learns a decision function between the true and estimated posteriors, that can then be used to quickly diagnose the SBI posterior for any new observation $\bx_o$. L-C2ST does so by comparing the predicted class probability to the chance level (i.e., $0.5$) of equal distributions. Any deviance from the chance level indicates a flawed inference.
Beyond P-P plots indicating overall too narrow or wide posteriors, plotting the classification probabilities over the posterior space can show us the regions where and how the posterior estimator should be improved.

\section{Experimental details on the examples}

\subsection{Gravitational wave simulator}
\label{sec:A4_1_grav_wave_details}
We use an implementation of the \texttt{pycbc} simulator \citep{pycbc_alex_nitz_2024_10473621} to generate synthetic observations and the training dataset.

The embedding network consists of a convolution layer with $16$ $1\times1$ kernels that expand the input signal by $8$. After this, a sequence of $13$ dilated convolutions is applied with kernel size of $2$. Each dilated convolution employs a dilation of factor $2^i$, with $i$ being the layer enumerator. Each dilated convolution is followed by a SELU activation function \citep{klambauer2017selfnormalizingneuralnetworks}. The overall architecture of the embedding network is inspired by previous publications in the domain for example \citep{hermans2022crisis}.

\subsection{Drift-diffusion model simulator}
\label{sec:A4_2_ddm_details}

This section provides technical details for practitioners implementing SBI on drift-diffusion models or similar trial-based data with mixed discrete-continuous observations.

\paragraph{Model specification.}
We used a five-parameter DDM variant with collapsing boundaries \citep{hawkins2015revisiting}. The parameters and their prior ranges were:
\begin{itemize}
    \item Drift rate $v \in [-2.5, 2.5]$: evidence accumulation rate
    \item Initial boundary separation $a \in [0.25, 1.0]$: decision threshold at $t=0$
    \item Starting point $w \in [-0.25, 0.25]$: bias toward upper/lower boundary
    \item Non-decision time $\tau \in [0.05, 0.95]$: sensory/motor delays
    \item Collapse rate $\gamma \in [-1.0, -0.1]$: boundary collapse speed (negative values indicate collapsing boundaries)
\end{itemize}
All parameters used uniform priors over these ranges, chosen based on typical values in the psychophysics literature.

\paragraph{Neural likelihood estimation with mixed data.}
Since DDM data contains both discrete choices (binary) and continuous reaction times, we employed Mixed Neural Likelihood Estimation (MNLE) \citep{boelts2022flexible}, a variant specifically designed for mixed discrete-continuous data. MNLE uses neural spline flow (NSF) \cite{durkan2019neural} for the continuous components and categorical distributions for discrete components, jointly modeling their dependencies. This architecture naturally handles the correlation between choices and reaction times in DDM data.

\paragraph{Training details.}
We trained the MNLE network on 200,000 single-trial simulations. Crucially, training on single trials enables efficient inference for datasets with varying numbers of trials without retraining. The network was trained using the default hyperparameters in the \texttt{sbi} package: Early stopping if the validation loss did not decrease for 20 epochs, batch size 200, Adam optimizer with learning rate $5 \times 10^{-4}$, and two NSF transforms with five spline bins, 2 layers, 50 hidden units each.

\paragraph{Inference procedure.}
For inference on multi-trial data, we computed the total log-likelihood by summing the single-trial log-likelihoods from the trained network. We then used MCMC (specifically, slice sampling with 100 chains) to sample from the posterior. Key MCMC settings included:
\begin{itemize}
    \item 1,000 warmup steps per chain
    \item Thinning factor of 2 to reduce autocorrelation
    \item Initialization via Sequential Importance Resampling (SIR) for better chain mixing
\end{itemize}
These settings balanced computational efficiency with sampling quality, particularly important when running diagnostics requiring hundreds of posterior inferences.

\paragraph{Computational considerations.}
The DDM simulator was parallelized across 10 workers to accelerate data generation. On a standard workstation, the training took approximately 30 minutes, while MCMC inference for a single 100-trial dataset required $\sim$2 minutes. The most computationally intensive step was the calibration diagnostics (SBC), which required 200 separate MCMC runs, taking $\sim$7 hours total.

\subsection{Pyloric network simulator from neuroscience}
\label{sec:A4_3_pyloric_details}

\paragraph{Simulator details.}
The simulator and 15 summary statistics are described in \citet{prinz2004similar}. \citet{gonccalves2020training} introduced three additional summary statistics. We used all 18 summary statistics. We obtained experimental data from \citet{haddad2018circuit}.

To deal with outliers, we capped the network cycle duration to 3000\,ms, the burst durations to 500\,ms, the duty cycles and start phases to 1, the burst gaps to 1000\,ms, the phase gaps to 1, and the time of extended high voltages to 10\,ms. We replaced undefined summary statistics with the same values (with the exception of the time of extended high voltages, which we replaced with 100\,ms).

\paragraph{Training details.}
We performed 3 million simulations. As the generative model, we used a neural spline flow with 10 transforms, each consisting of two hidden layers with 100 hidden neurons. We used a batchsize of 4096 and performed early stopping when the validation loss did not decrease for 30 epochs. Other parameters were the default values of the \texttt{sbi} package. We trained an ensemble of five such models (each with the same architecture, but with different random seeds for neural network initialization and training data shuffling) and averaged their posterior predictions (as implemented in the \texttt{sbi} toolbox).

\paragraph{Diagnostic details.}
We performed SBC and expected coverage based on a calibration set of 300 simulations. For L-C2ST, we used a calibration set of 20k simulations. As neural network for L-C2ST, we used the default implementation in the \texttt{sbi} toolbox \citep{boelts2024sbi}.

\end{document}